\documentclass[10pt,twocolumn,letterpaper,dvipsnames,svgnames,x11names]{article}

\usepackage{cvpr}
\usepackage{times}
\usepackage{epsfig}
\usepackage{graphicx}
\usepackage{amsmath}
\usepackage{amssymb}

\usepackage[utf8]{inputenc} 
\usepackage[T1]{fontenc}    
\usepackage{url}            
\usepackage{booktabs}       
\usepackage{amsfonts}       
\usepackage{mathtools}
\usepackage{nicefrac}       
\usepackage{microtype}      
\usepackage{tikz}
\usepackage{pgfplots}       
\usepackage{paralist}       
\usepackage{array}
\usepackage{subcaption}
\usepackage{makecell}
\usepackage[english,status=draft,innerlayout=]{fixme}
\usepackage{multirow}
\usepackage{xcolor}
\usepackage{colortbl}
\usepackage[acronym,smallcaps,nowarn,section,nogroupskip,nonumberlist]{glossaries}
\usepackage{setspace}

\usepackage[pagebackref=true,breaklinks=true,letterpaper=true,colorlinks,bookmarks=false,linkcolor={blue!80!black}, citecolor={blue!50!green}, urlcolor={red!70!gray}]{hyperref}
\usepackage[capitalise]{cleveref} 

\graphicspath{{images/}}
\usetikzlibrary{graphs, bayesnet, shapes, shapes.symbols, patterns, arrows, shadows}
\tikzset{latent/.append style={fill=none}}
\tikzset{const/.append style={inner sep=2pt,node distance=0.4}}
\tikzset{det/.style={const,draw,minimum size=18pt,node distance=0.6}}
\tikzset{>={stealth}}
\tikzstyle{arrow} = [->,>=stealth]
\tikzstyle{plate caption} = [caption, node distance=0, inner sep=0pt,
below left=0em and 0em of #1.south east] %
\crefformat{section}{\S#2#1#3}
\crefname{figure}{Fig.}{Figs.}
\crefname{table}{Tab.}{Tabs.}
\glsdisablehyper{}
\newacronym{HCI}{hci}{human-computer interaction}
\newacronym{CNN}{cnn}{convolutional neural network}
\newacronym{VAE}{vae}{variational auto-encoder}
\newacronym{CVAE}{cvae}{conditional variational auto-encoder}
\newacronym{KL}{kl}{{K}ullback-{L}eibler}
\newacronym{DNN}{dnn}{deep neural network}
\newacronym{SGD}{sgd}{stochastic gradient descent}
\newacronym{ELBO}{elbo}{evidence lower bound}
\newacronym{CE}{ce}{cross entropy}
\newacronym{VQA}{vqa}{visual question-answering}
\newacronym{1VD}{1vd}{one-way visual dialogue}
\newacronym{GVD}{gvd}{generative visual dialogue}
\newacronym{G2VD}{g2vd}{generative two-way visual dialogue}
\newacronym{2VD}{2vd}{two-way visual dialogue}
\newacronym{SVQA}{sqva}{Sequential Visual Question Answering}
\newacronym{AMT}{amt}{Amazon Mechanical Turk}

\fxusetheme{color}
\fxuselayouts{nomargin,inline}
\FXRegisterAuthor{sid}{esid}{sid}
\FXRegisterAuthor{dm}{edm}{daniela}
\FXRegisterAuthor{pd}{epd}{puneet}

\newcommand{\numberthis}{\refstepcounter{equation}\tag{\theequation}} \newcommand{\posref}[1]{{\bfseries #1}}

\setdefaultleftmargin{2ex}{2ex}{}{}{}{}
\newcolumntype{T}{>{\raggedright\let\newline\\\arraybackslash\hspace{0pt}}m{1.1cm}}
\newcolumntype{S}[1]{>{\raggedright\let\newline\\\arraybackslash\hspace{0pt}\columncolor{#1}}m{0.8cm}}

\makeatletter
\renewcommand{\paragraph}{%
  \@startsection{paragraph}{4}%
  {\z@}{3pt \@plus 2pt \@minus 1pt}{-2em}%
  {\normalfont\normalsize\bfseries}%
}
\makeatother

\usepackage{scalerel}
\usepackage{bm}
\usepackage{mleftright}
\newcommand\scalesym[2]{\hstretch{#1}{\vstretch{#1}{#2}}}

\DeclareMathOperator{\E}{{}\mathbb{E}}

\DeclareMathOperator{\DKL}{{}\mathbb{D}_{\text{\scalebox{0.75}{KL}}}}

\renewcommand{\to}{\ensuremath{\rightarrow}}              

\newcommand{\given}{\mid}
\renewcommand{\~}{\,\scalesym{0.7}{\thicksim}\,}          
\newcommand{\grpP}[1]{\ensuremath{\mleft( #1 \mright)}}   
\newcommand{\grpB}[1]{\ensuremath{\mleft\{ #1 \mright\}}} 
\newcommand{\grpS}[1]{\ensuremath{\mleft[ #1 \mright]}}   
\newcommand{\grpA}[1]{\ensuremath{\mleft< #1 \mright>}}   
\newcommand{\fnP}[2]{\ensuremath{#1 \grpP{#2}}}           
\newcommand{\fnS}[2]{\ensuremath{#1 \grpS{#2}}}           
\newcommand{\Ex}[2][]{\fnS{\E_{#1}}{#2}}                  
\newcommand{\KL}[2]{\fnP{\DKL}{#1 \,\|\; #2}}             

\newcommand{\p}[1]{\fnP{p_{\theta}}{#1}}
\newcommand{\q}[1]{\fnP{q_{\phi}}{#1}}
\newcommand{\e}[1]{\mathring{#1}}
\newcommand{\x}{\bm{x}}
\newcommand{\z}{\bm{z}}
\newcommand{\C}{\bm{y}}
\newcommand{\capt}{\bm{c}}
\newcommand{\img}{\bm{i}}
\newcommand{\ques}{\bm{q}}
\newcommand{\ans}{\bm{a}}
\newcommand{\dial}[1][]{\bm{d}_{#1}}
\newcommand{\ctx}{\bm{h}}

\newcommand{\fd}{\textsc{FlipDial}}
\newcommand{\pyt}{\textsc{PyTorch}}

\newcommand{\latent}{\mathbf{z}}

\newcommand{\mup}{\boldsymbol{\mu}_p}
\newcommand{\logvarp}{\log\boldsymbol{\sigma}^2_p}
\newcommand{\varp}{\boldsymbol{\sigma}^2_p}
\newcommand{\stddevp}{\boldsymbol{\sigma}_p}
\newcommand{\sigmap}{\boldsymbol{\sigma}^2_p}
\newcommand{\muq}{\boldsymbol{\mu}_q}
\newcommand{\logvarq}{\log\boldsymbol{\sigma}^2_q}
\newcommand{\varq}{\boldsymbol{\sigma}^2_q}
\newcommand{\stddevq}{\boldsymbol{\sigma}_q}
\newcommand{\sigmaq}{\boldsymbol{\sigma}^2_q}

\newcommand{\gaussianp}{\mathcal{N}(\latent; \mup, \varp)}
\newcommand{\gaussianq}{\mathcal{N}(\latent; \muq, \varq)}

\newcommand{\pl}{\texttt{+}}
\newcommand{\mn}{\texttt{-}}

\newcommand{\modx}{{\bf A}}
\newcommand{\modz}{{\bf B}}
\newcommand{\modzar}{{\(\textbf{B}_{\textbf{AR}}\)}}

\newcommand{\gtQA}{\(\dial\)--\(\ques\ans\)}
\newcommand{\predA}{\(\dial\)--\(\ques\hat{\ans}\)}
\newcommand{\predQA}{\(\dial\)--\(\hat{\ques}\hat{\ans}\)}

\newcommand{\pad}{{\small\texttt{PAD}}}

\cvprfinalcopy 

\def\cvprPaperID{3275} 

\ifcvprfinal\fi
\begin{document}

\title{\fd: A Generative Model for Two-Way Visual Dialogue}


\author{%
  Daniela Massiceti\\
  University of Oxford, UK\\
  {\tt\small daniela@robots.ox.ac.uk}
  \and
  N. Siddharth\\
  University of Oxford, UK\\
  {\tt\small nsid@robots.ox.ac.uk}
  \and
  Puneet K. Dokania\\
  University of Oxford, UK\\
  {\tt\small puneet@robots.ox.ac.uk}
  \and
  Philip H.S. Torr\\
  University of Oxford, UK\\
  {\tt\small phst@robots.ox.ac.uk}
}

\twocolumn[{
    \renewcommand\twocolumn[1][]{#1}
    \maketitle
    \centering
    \vspace*{-5ex}
    \includegraphics[width=0.99\textwidth]{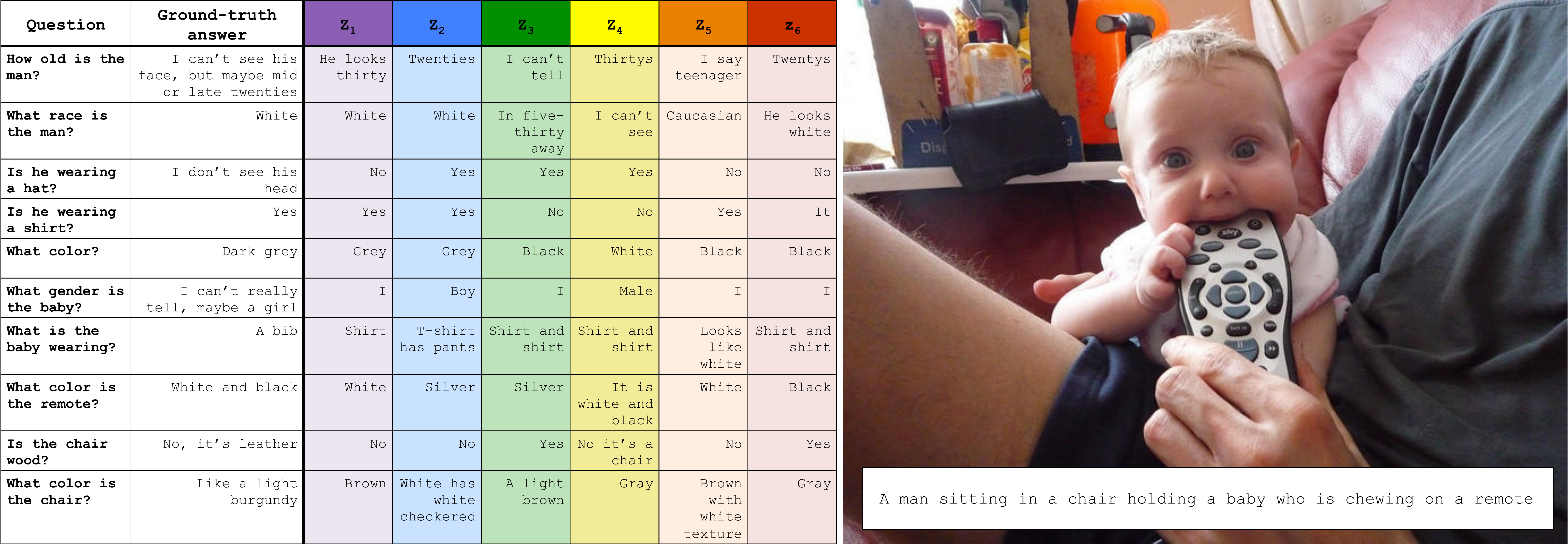}
    \vspace*{-1ex}
    \captionof{figure}{
      Diverse answers generated by \fd{} in the one-way visual dialogue (\acrshort{1VD}) task. For a given time step (row), each column shows a \emph{generated} answer to the current question. Answers are obtained by decoding a latent \(\z_i\) sampled from the conditional prior -- with conditions being the image, caption and dialogue history up until that time step.
    }
    \label{fig:gen-vd-hook}
    \vspace*{4ex}
}]

\begin{abstract}
  \vspace*{-1ex}
  We present \fd, a generative model for Visual Dialogue that simultaneously plays the role of both participants in a visually-grounded dialogue.
  Given context in the form of an image and an associated caption summarising the contents of the image, \fd{} learns \emph{both} to answer questions and put forward questions, capable of generating entire sequences of dialogue (question-answer pairs) which are diverse and relevant to the image.
  To do this, \fd{} relies on a simple but surprisingly powerful idea: it uses convolutional neural networks (CNNs) to encode entire dialogues directly, implicitly capturing dialogue context, and conditional VAEs to learn the generative model.
  \fd{} outperforms the state-of-the-art model in the sequential answering task (\acrshort{1VD}) on the \textit{VisDial} dataset by 5 points in Mean Rank using the generated answers.
  We are the first to extend this paradigm to full \gls{2VD}, where our model is capable of generating both questions {\em and} answers in sequence based on a visual input, for which we propose a set of novel evaluation measures and metrics.
  \vspace*{-1.5ex}
\end{abstract}

\vfill
\glsreset{1VD}
\glsreset{2VD}
\section{Introduction}
\label{sec:introduction}
\vspace*{-1ex}

A fundamental characteristic of a good \gls{HCI} system is its ability to effectively acquire \emph{and} disseminate knowledge about the tasks and environments in which it is involved.
A particular subclass of such systems, natural-language-driven conversational agents such as \emph{Alexa} and \emph{Siri}, have seen great success in a number of well-defined language-driven tasks.
Even such widely adopted systems suffer, however, when exposed to less circumscribed, more free-form situations.
Ultimately, an implicit requirement for the wide-scale success of such systems is the effective understanding of the environments and goals of the user -- an exceedingly difficult problem in the general case as it involves getting to grips with a variety of sub-problems (semantics, grounding, long-range dependencies) each of which are extremely difficult problems in themselves.
One avenue to ameliorate such issues is the incorporation of \emph{visual} context to help explicitly ground the language used -- providing a domain in which knowledge can be anchored and extracted from.
Conversely, this also provides a way in which language can be used to characterise visual information in richer terms, for example with sentences describing salient features in the image (referred to as ``captioning'')~\cite{Johnson_2016_CVPR,karpathy2015deep}.

In recent years, there has been considerable interest in visually-guided language generation in the form of \gls{VQA}~\cite{Antol_VQA} and subsequently visual dialogue~\cite{Das_VisualDialog}, both involving the task of \emph{answering} questions in the context of an image.
In the particular case of visual dialogue, along with the image, previously seen questions and answers (i.e.\ the dialogue history) are also accepted, and a relevant answer at the current time produced. We refer to this one-sided or answer-only form of visual dialogue as \gls{1VD}.
Inspired by these models and aiming to extend their capabilities, we establish the task of \gls{2VD} whereby an agent must be capable of acting as both the questioner and the answerer.

Our motivation for this is simple -- AI agents need to be able to both ask questions \emph{and} answer them, often interchangeably, rather do either one exclusively.
For example, a vision-based home-assistant (e.g.\ Amazon's \textit{Alexa}) may need to ask questions based on her visual input (``There is no toilet paper left. Would you like me to order more?'') but may also need to answer questions asked by humans (``Did you order the two-ply toilet paper?'').
The same question-answer capability is true for other applications.
For example, with aids for the visually-impaired, a user may need the answer to ``Where is the tea and kettle?'', but the system may equally need to query ``Are you looking for an Earl Grey or Rooibos teabag?'' to resolve potential ambiguities.

We take one step toward this broad research goal with \fd{}, a generative model capable of both \gls{1VD} and \gls{2VD}. The generative aspect of our model is served by using the \gls{CVAE}, a framework for learning deep conditional generative models while simultaneously amortising the cost of inference in such models over the dataset~\cite{Kingma_2014auto, Sohn_2015learning}.
Furthermore, inspired by the recent success of \glspl{CNN} in language generation and prediction tasks~\cite{hu2014convolutional,kalchbrennerconvolutional,pham2016convolutional}, we explore the use of \glspl{CNN} on sequences of sequences (i.e.\ a dialogue) to \emph{implicitly} capture all sequential dependences through the model.
Demonstrating the surprising effectiveness of this approach, we show sets of sensible and diverse answer generations for the \gls{1VD} task in \cref{fig:gen-vd-hook}.

We here provide a brief treatment of works related to visual dialogue. We reserve a
thorough comparison to Das et.al.~\cite{Das_VisualDialog} for \cref{sec:eval}, noting here that our fully-generative convolutional extension of their
model outperforms their state-of-the-art results on the answering of sequential visual-based questions (\gls{1VD}).
In another work, Das et.al.~\cite{das2017learning} present a Reinforcement Learning based
model to do \gls{1VD}, where they instantiate two separate agents, one
each for questioning and answering.
Crucially, the two agents are given \emph{different} information -- with one
(QBot) given the caption, and the other (ABot) given the image.
While this sets up the interesting task of performing image retrieval from
natural-language descriptions, it is also fundamentally different from having a
single agent perform both roles.
Jain et.al.~\cite{jain2017creativity} explore a complementary task to
\gls{VQA}~\cite{Antol_VQA} where the goal is instead to generate a (diverse) set of
relevant \emph{questions} given an image.
In their case, however, there is no dependence on a history of questions and
answers.
%
%
Finally, we note that Zhao et.al.~\cite{zhao2017learning} employ a similar
model structure to ours, using a \gls{CVAE} to model dialogue, but condition their model
on discourse-based constraints for a purely linguistic (rather than visuo-linguistic) dataset.
The tasks we target, our architectural differences (\glspl{CNN}), and the dataset and metrics we employ are distinct.

\noindent
Our primary contributions in this work are therefore:
\begin{compactitem}
\item A fully-generative, convolutional framework for visual dialogue that outperforms state-of-the-art models on sequential question answering (\gls{1VD}) using the generated answers, and establishes a baseline in the challenging two-way visual dialogue task (\acrshort{2VD}).
\item Evaluation using the \emph{predicted} (not ground-truth) dialogue -- essential for real-world conversational agents.
\item Novel evaluation metrics for generative models of two-way visual dialogue to quantify answer-generation quality, question relevance, and the models's generative capacity.
\end{compactitem}

\section{Preliminaries}
\label{sec:prelim}
Here we present a brief treatment of the preliminaries for deep generative models -- a conglomerate of deep neural networks and generative models.
In particular, we discuss the \gls{VAE}~\cite{Kingma_2014auto} which given a dataset
\(\mathcal{X}\) with elements \(\x \in \mathcal{X}\), simultaneously learns
\begin{inparaenum}[i)]
\item a variational approximation~\(\q{\z \given \x}\)\footnote{Following the literature, the terms recognition model or inference network may also be used to refer to the posterior variational approximation.} to the unknown posterior distribution~\(\p{\z \given \x}\) for latent variable~\(\z\), and
\item a generative model~\(\p{\x, \z}\) over data and latent variables.
\end{inparaenum}
These are both highly attractive prospects as the ability to approximate the posterior distribution helps \emph{amortise} inference for any given data point~\(\x\) over the entire dataset~\(\mathcal{X}\), and learning a generative model helps effectively capture the underlying abstractions in the data.
Learning in this model is achieved through a unified objective, involving the marginal likelihood (or \emph{evidence}) of the data, namely:
\begin{align*}
  \log \p{\x}
  &= \KL{\q{\z \given \x}}{\p{\z \given \x}}\\
  &\quad+ \Ex[\q{\z \given \x}]{\log \p{\x, \z} - \log \q{\z \given \x}}\\
  &\ge \Ex[\q{\z | \x}]{\log \p{\x | \z}} - \KL{\q{\z | \x}\!}{\!\p{\z}}
  \numberthis
  \label{eq:elbo}
\end{align*}
The unknown true posterior~\(\p{\z \given \x}\) in the first \gls{KL} divergence is intractable to compute making the objective difficult to optimise directly. Rather a lower-bound of the marginal log-likelihood \(\log \p{\x}\), referred to as the \gls{ELBO}, is maximised instead.

By introducing a condition variable~\(\C\), we capture a {\em conditional} posterior approximation \(\q{\z \given \x, \C}\) and a {\em conditional} generative model \(\p{\x, \z \given \C}\), thus deriving the \gls{CVAE}~\cite{Sohn_2015learning}. Similar to \cref{eq:elbo}, the conditional \gls{ELBO} is:
\begin{align*}
  \log \p{\x \given \C}
  &\ge \Ex[\q{\z \given \x, \C}]{\log \p{\x \given \z, \C}}\\
  &\quad- \KL{\q{\z \given \x, \C}}{\p{\z \given \C}}
  \numberthis
  \label{eq:c-elbo}
\end{align*}
where the first term is referred to as the reconstruction or negative \gls{CE} term, and the second, the regularisation or \gls{KL} divergence term. Here too, similar to the \gls{VAE}, \(\q{\z \given \x, \C}\) and \(\p{\z \given \C}\) are typically taken to be isotropic multivariate Gaussian distributions, whose parameters~\((\muq, \sigmaq)\) and~\((\mup, \sigmap)\) are provided by \glspl{DNN} with parameters~\(\phi\) and~\(\theta\), respectively.
The generative model likelihood~\(\p{\x \given \z, \C}\), whose form varies depending on the data type -- Gaussian or Laplace for images and Categorical for language models -- is also parametrised similarly.
In this work, we employ the \gls{CVAE} model for the task of eliciting dialogue \emph{given} contextual information from vision (images) and language (captions).

\glsreset{1VD}
\glsreset{2VD}
\section{Generative Models for Visual Dialogue}
\label{sec:vd-gen}

In applying deep generative models to visual dialogue, we begin by characterising a preliminary step toward it, \gls{VQA}.
In \gls{VQA}, the goal is to answer a single question in the context of a visual cue, typically an image.
The primary goal for such a model is to ensure that the elicited answer conforms to a stronger notion of relevance than simply answering the given question -- it must also relate to the visual cue provided.
This notion can be extended to \gls{1VD} which we define as the task of answering a \emph{sequence} of questions contextualised by an image (and a short caption describing its contents), similar to~\cite{Das_VisualDialog}.
Being able to exclusively answer questions, however, is not fully encompassing of true conversational agents. We therefore extend \gls{1VD} to the more general and realistic task of \gls{2VD}. Here the model must elicit not just answers given questions, but questions given answers as well -- generating \emph{both} components of a dialogue, contextualised by the given image and caption.
Generative \gls{1VD} and \gls{2VD} models introduce stochasticity in the latent representations.

As such, we begin by characterising our generative approach to \gls{2VD} using a \gls{CVAE}.
For a given image~\(\img\) and associated caption~\(\capt\), we define a dialogue as a sequence of question-answer pairs~\(\dial[1:T] = \grpA{(\ques_t, \ans_t)}^T_{t=1}\), simply denoted~\(\dial\) when sequence indexing is unnecessary.
Additionally, we denote a dialogue context~\(\ctx\). When indexed by step as~\(\ctx_t\), it captures the dialogue subsequence~\(\dial[1:t]\).

\begin{figure}[h!]
  \centering
  \begin{tikzpicture}[node distance=0.7em, auto, thick]
    \node[const] (img) {\(\img\)};
    \node[const, right=of img] (capt) {\(\capt\)};
    \node[const, right=of capt] (ctx) {\(\ctx\)};
    \node[obs, below right=of ctx] (dial) {\(\dial\)};
    \node[latent, left=of dial] (z) {\(\z\)};
    \edge{dial,img,capt,ctx} {z};
  \end{tikzpicture}
  \hspace*{3ex}
  \begin{tikzpicture}[node distance=0.7em, auto, thick]
    \node[const] (img) {\(\img\)};
    \node[const, right=of img] (capt) {\(\capt\)};
    \node[const, right=of capt] (ctx) {\(\ctx\)};
    \node[latent, below left=of img] (z) {\(\z\)};
    \node[obs, right=of z] (dial) {\(\dial\)};
    \edge{z,img,capt,ctx} {dial};
  \end{tikzpicture}
  \caption{%
    \posref{Left:} Conditional recognition model and \posref{Right:} conditional generative model for \gls{2VD}.
  }
  \label{fig:vd-gm}
\end{figure}
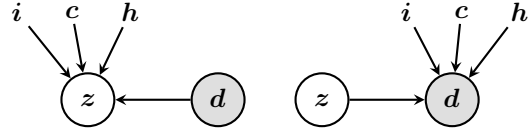

With this formalisation, we characterise a generative model for \gls{2VD} under latent variable~\(\z\) as \(  \p{\dial, \z \given \img, \capt, \ctx} =  \p{\dial \given \z, \img, \capt, \ctx}\; \p{\z \given \img, \capt, \ctx} \), with the corresponding recognition model defined as \(\q{\z \given \dial, \img, \capt, \ctx}\).
Note that with relation to \cref{eq:c-elbo}, data~\(\x\) is dialogue~\(\dial\)
and the condition variable is~\(\C = \grpB{\img, \capt, \ctx}\), giving:
\begin{align*}
  &\log \p{\dial \given \img, \capt, \ctx}\\
  &\ge \Ex[\q{\z \given \dial, \img, \capt, \ctx}]{\log \p{\dial \given \z, \img, \capt, \ctx}}\\
  &\quad- \KL{\q{\z \given \dial, \img, \capt, \ctx}}{\p{\z \given \img, \capt, \ctx}},
  \numberthis
  \label{eq:vd-elbo}
\end{align*}
with the graphical model structures shown in \cref{fig:vd-gm}.

The formulation in \cref{eq:vd-elbo} is general enough to be applied to single question-answering (\gls{VQA}) all the way to full two-way dialogue generation (\gls{2VD}).
Taking a step back from generative \gls{2VD}, we can re-frame the formulation for generative \gls{1VD} (i.e.\ sequential answer generation) by considering the generated component to be the answer to a particular question at step~\(t\), given context from the image, caption and the sequence of previous question-answers.
Simply put, this corresponds to the data~\(\x\) being the answer~\(\ans_t\), conditioned on the image, its caption, the dialogue history to \(t\)-1, and the current question, or \(\C = \grpB{\img, \capt, \ctx_{t-1}, \ques_t}\).
For simplicity, we denote a compound context as~\(\ctx^{\pl}_t = \grpA{\ctx_{t-1}, \ques_t}\) and reformulate \cref{eq:vd-elbo} for \gls{1VD} as:
\begin{align*}
  &\log \p{\dial \given \img, \capt, \ctx} = \sum_{t=1}^T \log \p{\ans_t \given \img, \capt, \ctx^{\pl}_t},\\[-2ex]
  &\log \p{\ans_t \given \img, \capt, \ctx^{\pl}_t}\\
  &\ge \Ex[\q{\z \given \ans_t, \img, \capt, \ctx^{\pl}_t}]
          {\log \p{\ans_t \given \z, \img, \capt, \ctx^{\pl}_t}}\\
  &\quad- \KL{\q{\z \given \ans_t, \img, \capt, \ctx^{\pl}_t}}
             {\p{\z \given \img, \capt, \ctx^{\pl}_t}},
  \numberthis
  \label{eq:svqa-g-elbo}
\end{align*}
with the graphical model structures shown in \cref{fig:svqa-g-gm}.
\begin{figure}[h!]
  \centering
  \begin{tikzpicture}[node distance=0.7em, auto, thick]
    \node[const] (img) {\(\img\)};
    \node[const, below=of img] (capt) {\(\capt\)};
    \node[latent, right=of capt] (z) {\(\z\)};
    \node[const, above=of z] (ctx) {\(\ctx^{\pl}_t\)};
    \node[obs, right=1.5em of z] (dial) {\(\ans_t\)};
    \edge{dial,img,capt,ctx} {z};
    \plate [inner xsep=8pt]{r} {(ctx)(dial)} {\rule[1.2em]{0pt}{0pt}$T$};
  \end{tikzpicture}
  \hspace*{3ex}
  \begin{tikzpicture}[node distance=0.7em, auto, thick]
    \node[const] (img) {\(\img\)};
    \node[const, below=of img] (capt) {\(\capt\)};
    \node[obs, left=of capt] (dial) {\(\ans_t\)};
    \node[latent, left=1.5em of dial] (z) {\(\z\)};
    \node[const, above=of dial] (ctx) {\(\ctx^{\pl}_t\)};
    \edge{z,img,capt,ctx} {dial};
    \plate [inner xsep=8pt]{r} {(z)(ctx)} {\rule[1.2em]{0pt}{0pt}$T$};
  \end{tikzpicture}
  \caption{%
    \posref{Left:} Conditional recognition model and \posref{Right:} conditional generative model for \gls{1VD}.
  }
  \label{fig:svqa-g-gm}
\end{figure}

Our baseline~\cite{Das_VisualDialog} for the \gls{1VD} model can also be represented in our formulation by taking the variational posterior and generative prior to be conditional Dirac-Delta distributions. That is, \(\q{\z \given \ans_t, \img, \capt, \ctx^{\pl}_t} = \p{\z \given \img, \capt, \ctx^{\pl}_t} = \delta(\z \given \img, \capt, \ctx^{\pl}_t)\).
This transforms the objective from \cref{eq:svqa-g-elbo} by
\begin{inparaenum}[a)]
\item replacing the expectation of the log-likelihood over the recognition model by an evaluation of the log-likelihood for a \emph{single} encoding (one that satisfies the Dirac-Delta), and
\item ignoring the \(\DKL\) regulariser, which is trivially 0.
\end{inparaenum}
This computes the marginal likelihood directly as just the model likelihood~\(\log \p{\ans_t \given \z, \img, \capt, \ctx^{\pl}_t}\), where~\(\z \~ \delta(\z \given \img, \capt, \ctx^{\pl}_t)\).

Note that while such models can ``generate'' answers to questions by sampling from the likelihood function, we typically don't call them generative since they effectively make the encoding of the data and conditions fully deterministic.
We explore and demonstrate the benefit of a fully generative treatment of \gls{1VD} in \cref{sec:eval}.
It also follows trivially that the basic \gls{VQA} model (for single question-answering) itself can be obtained from this \gls{1VD} model by simply assuming there is no dialogue history (i.e.\ step length \(T = 1\)).

\subsection{``Colouring'' Visual Dialogue with Convolutions}
\label{sec:vd-conv}

\fd's convolutional formulation allows us to {\em implicitly} capture the sequential nature of sentences and sequences of sentences. Here we introduce how we encode questions, answers, and whole dialogues with \glspl{CNN}.

We begin by noting the prevalence of recurrent approaches (e.g.\ LSTM~\cite{Hochreiter1997}, GRU~\cite{chung2014empirical}) in modelling both visual dialogue and general dialogue to date~\cite{Das_VisualDialog,das2017learning,visdial_rl,jain2017creativity,zhao2017learning}.
Typically recurrence is employed at two levels -- at the lower level to sequentially generate the words of a sentence (a question or answer in the case of dialogue), and at a higher level to sequence these sentences together into a dialogue.

Recently however, there has been considerable interest in convolutional models of language~\cite{Bai_2018conv,hu2014convolutional,kalchbrennerconvolutional,pham2016convolutional}, which have shown to perform at least as well as recurrent models, if not better, on a number of different tasks.
They are also computationally more efficient, and typically suffer less from issues relating to exploding or vanishing gradients for which recurrent networks are known~\cite{pascanu2013difficulty}.

In modelling sentences with convolutions, the tokens (words) of the sentence are transformed into a stack of fixed-dimensional embeddings (e.g.\ using word2vec~\cite{mikolov2013distributed} or Glove~\cite{pennington2014glove}, or those learned for a specific task).
For a given sentence, say question~\(\ques_t\), this results in an embedding~\(\e{\ques_t} \in \mathbb{R}^{E \times L}\) for embedding size~\(E\) and sentence length~\(L\), where \(L\) can be bounded by the maximum sentence length in the corpus, with padding tokens employed where required.
This two-dimensional stack is essentially a single-channel `image' on which convolutions can be applied in the standard manner in order to encode the entire sentence.
Note this similarly applies to the answer~\(\ans_t\) and caption~\(\capt\), producing embedded~\(\e{\ans}_t\) and~\(\e{\capt}\), respectively.

We then extend this idea of viewing sentences as `images' to whole dialogues, producing a \emph{multi-channel} language embedding.
Here, the sequence of sentences itself can be seen as a stack of (a stack of) word embeddings~\(\e{\dial} \in \mathbb{R}^{E \times L \times 2T}\), where now the number of channels accounts for the number of questions and answers in the dialogue.
We refer to this process as ``colouring'' dialogue, by analogy to the most common meaning given to image channels -- colour.

Our primary motivation for adopting a convolutional approach here is to explore its efficacy in extending from simpler language tasks~\cite{hu2014convolutional, kalchbrennerconvolutional} to full visual dialogue. We hence instantiate the following models for \gls{1VD} and \gls{2VD}:
\begin{compactdesc}
\item [{\bf Answer} {[\gls{1VD}]}:] We employ the \gls{CVAE} formulation from \cref{eq:svqa-g-elbo,fig:svqa-g-gm} to iteratively generate answers, conditioned on the image, caption and current dialogue history.
\item[{\bf Block} {[\gls{1VD}, \gls{2VD}]}:] Using the \gls{CVAE} formulation from \cref{eq:vd-elbo,fig:vd-gm} we generate entire \emph{blocks} of dialogue directly (i.e.~\(\ctx=\emptyset\) since dialogue context is implicit rather than explicit). We allow the convolutional model to \emph{implicitly} supply the context instead. We consider this \gls{2VD}, although this block architecture can also generate iteratively, and can be evaluated on \gls{1VD} (see~\cref{sec:blockevalmethods}).
\item[{\bf Block Auto-Regressive} {[\gls{1VD}, \gls{2VD}]}:] We introduce an auto-regressive component to our generative model in the same sense as recent auto-regressive generative models for images~\cite{gulrajani2016pixelvae,van2016conditional}.
  We augment the {\bf Block} model by feeding its output through an auto-regressive ({\sc AR}) module which explicitly enforces sequentiality in the generation of the dialogue blocks.
  This effectively factorises the likelihood in \cref{eq:vd-elbo} as \(\p{\dial \given \z, \img, \capt, \ctx} = \p{\dial^1 \given \z, \img, \capt, \ctx} \prod^{N}_{n=2} \p{\dial^n \given \dial^{1:n-1}}\) where \(N\) is the number of {\sc AR} layers, and \(\dial^1\) is the (intermediate) output from the standard {\bf Block} model. Note, again \(\ctx=\emptyset\), and \(\dial^{n}\) refers to an entire dialogue at the \(n\)-th {\sc AR} layer (rather than the \(t\)-th dialogue exchange as is denoted by \(\dial[t]\)).
\end{compactdesc}

\vspace*{-2ex}
\section{Experiments}
\label{sec:experiments}
We present an extensive quantitative and qualitative analysis of our models' performance in both \gls{1VD}, which requires answering a sequence of image-contextualised questions, and full \gls{2VD}, where both questions \textit{and} answers must be generated given a specific visual context. Our proposed generative models are denoted as follows:
\noindent%
\begin{tabular}{>{\,\,\bfseries}r@{\,\,--\,\,}l@{}}
  \modx   & {\bf a}nswer architecture for \gls{1VD}\\
  \modz   & {\bf b}lock dialogue architecture for \gls{1VD} \& \gls{2VD} \\
  \modzar & auto-regressive extension of \modz{} for \gls{1VD} \& \gls{2VD}
\end{tabular}

\noindent%
\modx{} is a generative convolutional extension of our baseline~\cite{Das_VisualDialog} and is used to validate our methods against a standard benchmark in the \gls{1VD} task. \modz{} and \modzar, like \modx, are generative, but are extensions capable of doing full dialogue generation, a much more difficult task. Importantly, \modz{} and \modzar{} are flexible in that despite being trained to generate a block of questions \emph{and} answers (\(\ctx=\emptyset\)), they can be \emph{evaluated} iteratively for both \gls{1VD} and \gls{2VD} (see~\cref{sec:blockevalmethods}). We summarise the data and condition variables for all models in \cref{tab:methodsumm}. To evaluate performance on both tasks, we propose novel evaluation metrics which augment those of our baseline~\cite{Das_VisualDialog}. To the best of our knowledge, we are the first to report models that can generate both questions and answers given an image and caption, a necessary step toward a truly conversational agent. Our key results are:
\begin{compactitem}
\item We set state-of-the-art results in the \gls{1VD} task on the \textit{VisDial} dataset, improving the mean rank of the generated answers by \(5.66\) (\cref{tab:results_ansgen}, \(\mathcal{S}_{\textit{w2v}}\)) compared to Das~\etal~\cite{Das_VisualDialog}.
\item Our block models are able to generate both questions and answers, a more difficult but more realistic task (\gls{2VD}).
\item Since our models are generative, we are able to show highly diverse and plausible question and answer generations based on the provided visual context.
\end{compactitem}

\begin{table}
  \vspace*{3ex}                 
  \caption{Data (\(\x\)) and condition (\(\C\)) variables for models \modx{} and \modz{}/\modzar{} for \gls{1VD} and \gls{2VD}. Models \modz/\modzar{} can be evaluated as a block or iteratively (see \cref{sec:blockevalmethods}), accepting ground-truth (\(\ques / \ans\)) or predicted (\(\hat{\ques} / \hat{\ans}\)) dialogue history (see \cref{tab:blockevalmethods}). }
  \label{tab:methodsumm}
  \centering
  \scalebox{0.72}{%
  \begin{tabular}{@{}ccccccc@{}}
    \toprule
    Task & Model & \multicolumn{2}{c}{Train} & \multicolumn{2}{c}{Evaluate} & Eval method \\
    \cmidrule(lr){3-6}
    & & {\bf \(\x\)} & \bf \(\C\) & {\bf \(\x\)} & \bf \(\C\) &  \\
    \midrule
    \multirow{2}*{\gls{1VD}} & \modx & \(\ans_t\) & \(\img, \capt, \ctx^{\pl}_t\) &  \(\emptyset\) & \(\img, \capt, \ctx^{\pl}_t\) & \(-\) \\
    & \modz{}, \modzar{} & \(\dial\) & \(\img, \capt\) & \{\gtQA, \predA\} & \(\img, \capt\) & iterative \\
    \midrule
    \multirow{2}*{\gls{2VD}} & \multirow{2}*{\modz{}, \modzar{}} & \multirow{2}*{\(\dial\)} & \multirow{2}*{\(\img, \capt\)} & \(\emptyset\) & \multirow{2}*{\(\img, \capt\)} & block \\
    & & & &  \predQA & & iterative\\
    \bottomrule
  \end{tabular}}
\vspace*{-3ex}
\end{table}

\paragraph{Datasets:}%
We use the \textit{VisDial}~\cite{Das_VisualDialog} dataset (v0.9) which contains Microsoft COCO images each paired with a caption and a dialogue of 10 question-answer pairs. The train/test split is \(82,783/40,504\) images, respectively.

\paragraph{Baseline:}%
Das \etal~\cite{Das_VisualDialog}'s best model, \texttt{MN-QIH-G}, is a recurrent encoder-decoder architecture which encodes the image \( \img \), the current question \( \ques_t \) and the attention-weighted \textit{ground truth} dialogue history \( \dial[1:t-1]\). The output conditional likelihood distribution is then used to (token-wise) predict an answer. Our \modx{} model is a generative and convolutional extension, evaluated using existing ranking-based metrics~\cite{Das_VisualDialog} on the generated and candidate answers. We also (iteratively) evaluate our \modz/\modzar{} for \gls{1VD} as detailed in \cref{sec:blockevalmethods} (see \cref{tab:results_ansgen}).

\vspace*{-0.4ex}
\subsection{Network architectures and training}
\vspace*{-0.3ex}
\begin{figure}
  \centering
  \includegraphics[width=0.45\textwidth]{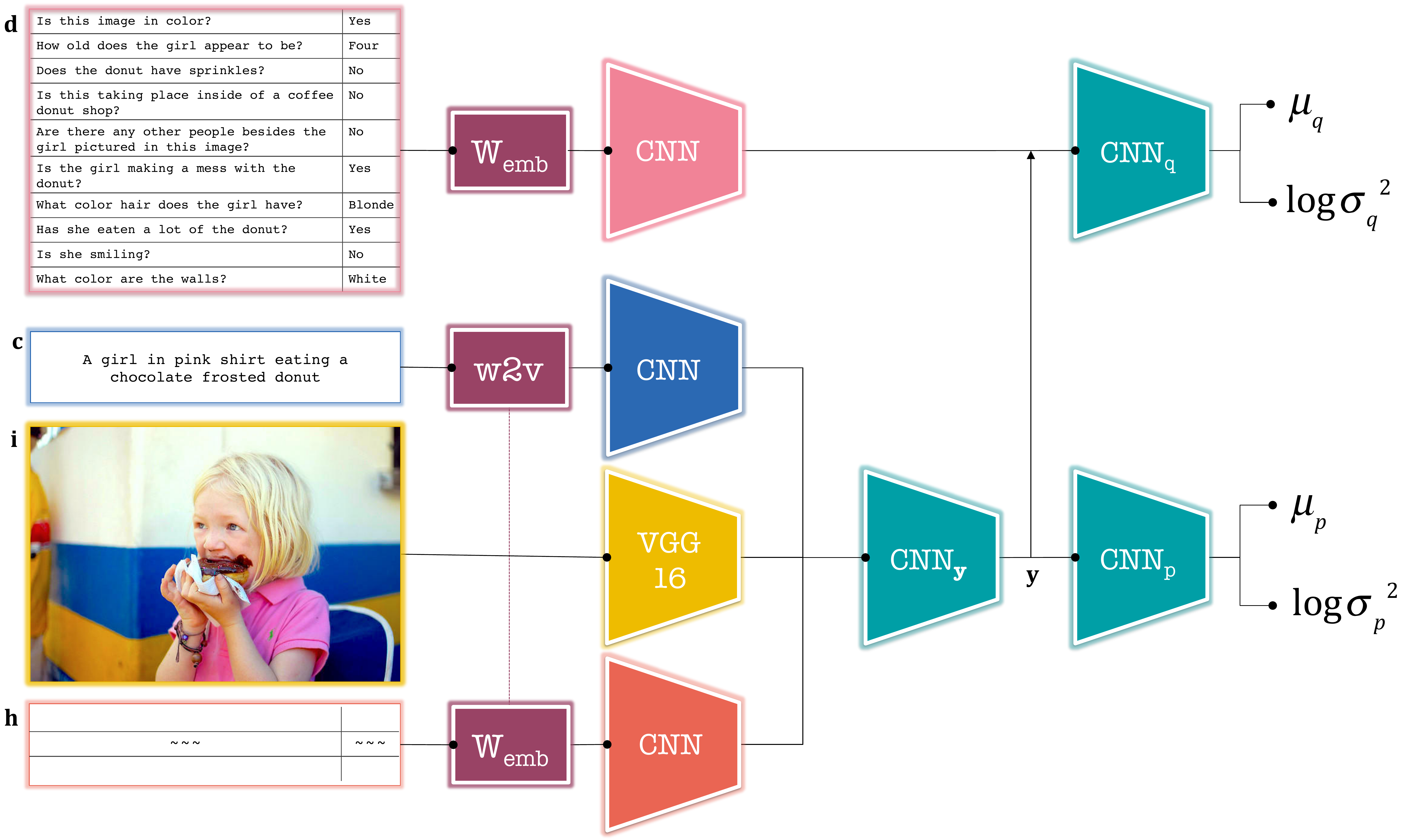}\\
  \includegraphics[width=0.45\textwidth]{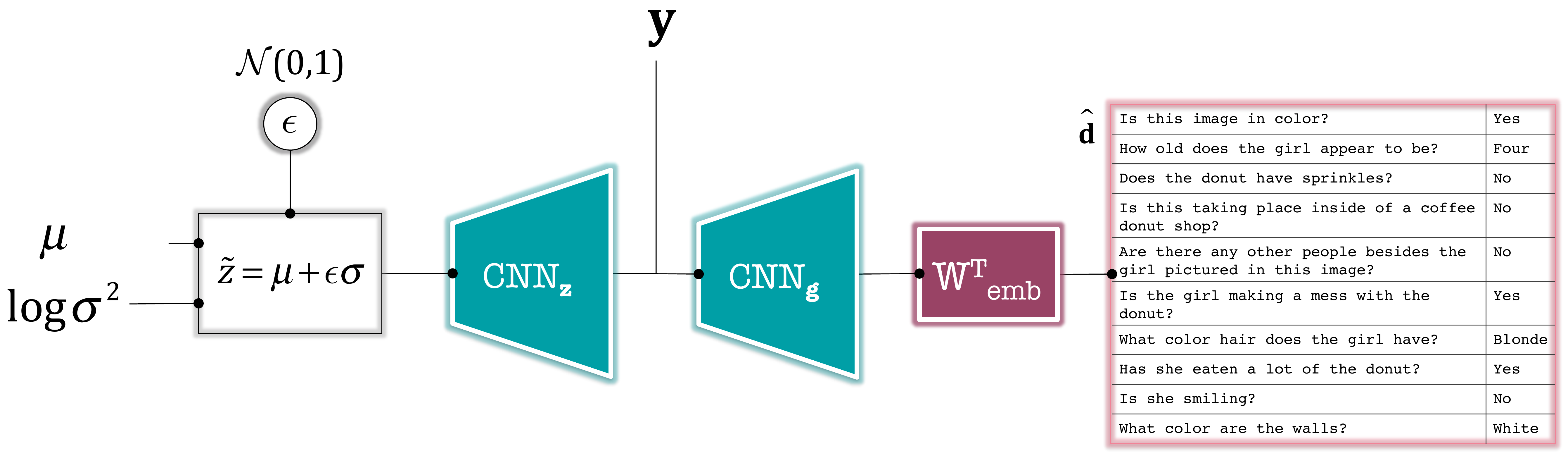}\\
  \includegraphics[width=0.45\textwidth]{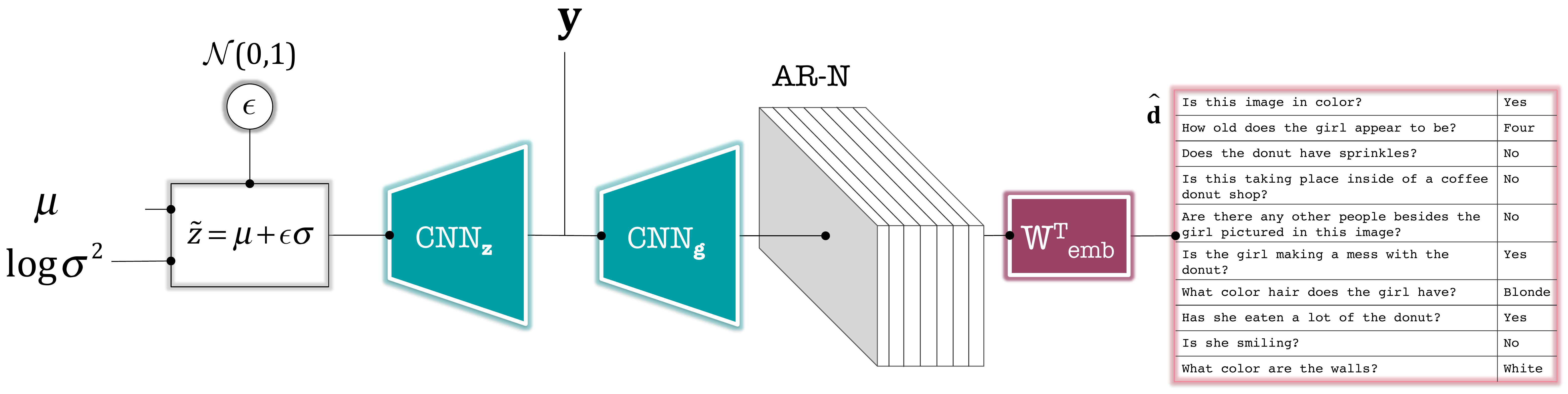}
  \vspace*{-1ex}
  \caption{Convolutional \posref{(top)} conditional encoder and prior architecture, \posref{(middle)} conditional decoder, and \posref{(bottom)} auto-regressive conditional decoder architectures, applying to both one- and two-way visual dialogue (\gls{1VD} and \gls{2VD}).}
  \label{fig:nw-arch}
\vspace*{-1ex}
\end{figure}

Following the \gls{CVAE} formulation (\cref{sec:vd-gen}) and its convolutional interpretation (\cref{sec:vd-conv}), all our  models (\modx, \modz{} and \modzar) have three core components: an encoder network, a prior network and a decoder network. \Cref{fig:nw-arch}~(top) shows the encoder and prior networks, and \cref{fig:nw-arch}~(middle, bottom) show the standard and auto-regressive decoder networks.

\paragraph{Prior network}
The prior neural network, parametrised by \(\theta\), takes as input the image \(\img\), the caption \(\capt\) and the dialogue context. Referring to Table~\ref{tab:methodsumm}, for model \modx, recall \(\C=\{\img, \capt, \ctx^{\pl}_t\}\) where the context \(\ctx^{\pl}_t\) is the dialogue history up to \(t\text{-}1\) and the current question \(\ques_t\). For models \modz/\modzar, \(\C=\{\img, \capt\}\) (note \(\ctx=\emptyset\)).
To obtain the image representation, we pass \(\img\) through \textit{VGG-16}~\cite{Simonyan_2014c} and extract the penultimate (\(4096\)-d) feature vector.
We pass caption \(\capt\) through a pre-trained \textit{word2vec}~\cite{mikolov2013distributed} module (we do not learn these word embeddings).
If \(\ctx \neq \emptyset\), we pass the one-hot encoding of each word through a {\em learnable} word embedding module and stack these embeddings as described in \cref{sec:vd-conv}. We encode these condition variables convolutionally to obtain \(\C\), and pass this through a convolutional block to obtain \(\mup\)  and \(\logvarp\), the parameters of the conditional prior \(\p{\z \given \C}\).

\paragraph{Encoder network} The encoder network, parametrised by \(\phi\), takes \(\x\) and the encoded condition \(\C\) (obtained from the prior network) as input.
For model \modx, \(\x=\ans_t\) while for \modz{}/\modzar, \(\x\!=\!\dial\!=\!\grpA{(\ques_t, \ans_t)}^T_{t=1}\).
In all models, \(\x\) is transformed through a word-embedding module into a single-channel answer `image' for \modx{}, or a multi-channel image of alternating questions and answers for \modz/\modzar. The embedded output is then combined with \(\C\) to obtain \(\muq\)  and \(\logvarq\), the parameters of the conditional latent posterior \(\q{\z \given \x, \C}\).

\paragraph{Decoder network} The decoder network 
takes as input a latent \(\z\) and the encoded condition \(\C\). 
%
%
%
The sample is transpose-convolved, combined with \(\C\) and further transformed to obtain an intermediate output volume of dimension \(E \times L \times M\), where \(E\) is the word embedding dimension, \(L\) is the maximum sentence length and \(M\) is the number of dialogue entries in \(\x\) (\(M=1\) for \modx{}, \(M=2T\) for \modz{} variants).
Following this, 
\modx{} and \modz{} employ a standard linear layer, projecting the \(E\) dimension to the vocabulary size \(V\) (\cref{fig:nw-arch}~(middle)), whereas \modzar{} employs an autoregressive module followed by this standard linear layer (\cref{fig:nw-arch}~(bottom)).
At train time, the \(V\)-dimensional output is \textit{softmax}ed and the \gls{CE} term of the \gls{ELBO} computed. At test time, the \(\mathop{argmax}\) of the output provides the predicted word index.
The weights of the encoder and prior's learnable word embedding module and the decoder's final linear layer are shared.

\paragraph{Autoregressive module}
Inspired by \textit{PixelCNN}~\cite{oord2016pixelrnn} which sequentially predicts image pixels, and similar to \cite{gulrajani2016pixelvae}, we apply \(N = \{8, 10\}\) size-preserving autoregressive layers to the intermediate output of model \modz{} (size \(E \times L \times 2T\)), and then project \(E\) to vocabulary size \(V\). Each layer employs masked convolutions, considering only `past' embeddings, sequentially predicting \(2T*L\) embeddings of size \(E\), enforcing sequentiality at both the sentence- and dialogue-level.

\paragraph{\gls{KL} annealing}
Motivated by \cite{bowman2015generating} in learning continuous latent embedding spaces for language, we employ \gls{KL} annealing in the loss objectives of \cref{eq:vd-elbo} and \cref{eq:svqa-g-elbo}.
We weight the \gls{KL} term by \(\alpha \in [0,1]\) linearly interpolated over 100 epochs, and then train for a further 50 epochs (\(\alpha=1\)).

\paragraph{Network and training hyper-parameters}
In embedding sentences, we pad to a maximum sequence length of \(L=64\) and use a word-embedding dimension of \(E=256\) (for \textit{word2vec}, \(E=300\)).
After pre-processing and filtering the vocabulary size is \(V=9710\) (see supplement for further details).
We use the Adam optimiser~\cite{Kingma_2014adam} with default parameters, a latent dimensionality of \(512\) and employ batch normalisation with momentum\(=0.001\) and learnable parameters.
For model \modx{} we use a batch size of \(200\), and \(40\) for \modz{}/\modzar.
We implement our pipeline using \pyt{}~\cite{pytorch}.

\begin{table}
  \vspace*{3ex}
  \caption{Iterative evaluation of \modz{}/\modzar{} for \gls{1VD} and \gls{2VD}. Under each condition, the input dialogue block is filled with ground-truth or predicted history (\(\ques/\ans\) or \(\hat{\ques}/\hat{\ans}\), respectively), while future entries are filled with the \pad{} token. }
  \label{tab:blockevalmethods}
  \vspace*{-1ex}
  \centering
  \scalebox{0.85}{%
  \begin{tabular}{@{}c@{\hspace*{5ex}}cc@{\hspace*{5ex}}c@{}}
    \toprule
    &  \multicolumn{2}{c}{\gls{1VD}} & \gls{2VD} \\
    \cmidrule{2-4}
    & \gtQA & \predA & \predQA \\
    \midrule
    \({}<t\) & (\(\ques, \ans\)) & (\(\ques, \hat{\ans}\)) & (\(\hat{\ques}, \hat{\ans}\))\\
    \({}=t\) & (\(\ques\), \pad{}) & (\(\ques\), \pad{}) & (\pad{}, \pad{}) \big/ (\(\hat{\ques}\), \pad{})\\

    \({}>t\) & (\pad{}, \pad{}) & (\pad{}, \pad{}) & (\pad{}, \pad{})\\
    \bottomrule
  \end{tabular}}
\vspace*{-3ex}
\end{table}

\subsection{Evaluation methods for block models}
\label{sec:blockevalmethods}
Although \modz{}/\modzar{} generate whole blocks of dialogue directly (\(\ctx=\emptyset\)), they can be evaluated iteratively, lending them to both \gls{1VD} and \gls{2VD} (see supplement for descriptions of generation/reconstruction pipelines).
\begin{compactitem}
\item {\bf Block evaluation [\gls{2VD}]}. The generation pipeline generates whole blocks of dialogue directly, conditioned on the image and caption (i.e.\ \(\x=\emptyset\) and \(\C=\{\img, \capt\}\) for \modz/\modzar{} evaluation in \cref{tab:methodsumm}). This is \gls{2VD} since the model must generate a coherent block of both questions {\em and} answers.
\item {\bf Iterative evaluation}. The reconstruction pipeline can generate dialogue items iteratively. At time \(t\), the input dialogue block is filled with zeros (\pad{} token) and the ground-truth/predicted dialogue history to \(<t\) is slotted in (see below and \cref{tab:blockevalmethods}). This future-padded block is then encoded with the condition inputs, and then reconstructed. The \(t\)-th dialogue item is extracted (whether an answer if \gls{1VD} or a question/answer if \gls{2VD}), and this is repeated \(T\) (for \gls{1VD}) or \(2T\) (for \gls{2VD}) times. Variations are:
  \begin{compactitem}
  \item \gtQA{} {\bf [\gls{1VD}]}. At time \(t\), the input dialogue block is filled with the history of {\em ground-truth} questions and answers up to~\(t\text{-}1\), along with the current ground-truth question. All future entries are padded -- equivalent to~\cite{Das_VisualDialog} using the ground-truth dialogue history.
  \item \predA{} {\bf [\gls{1VD}]}. Similar to \gtQA, except that the input block is filled with the history of ground-truth questions and {\em previously predicted} answers along with the current ground-truth question. This is a more realistic \gls{1VD}.
  \item \predQA{} {\bf [\gls{2VD}]}. The most challenging and realistic condition in which the input block is filled with the history of previously predicted questions {\em and} answers. 
  \end{compactitem}
\end{compactitem}

\subsection{Evaluation and Analysis}
\label{sec:eval}
We evaluate our \modx, \modz, and \modzar{} models on the \gls{1VD} and \gls{2VD} tasks. Under \gls{1VD}, we predict an answer with each time step, given an image, caption and the current dialogue history (\cref{sec:vd} and \cref{tab:results_ansgen}), while under \gls{2VD}, we predict both questions \emph{and} answers (\cref{sec:twowayvd} and \Cref{tab:results_modz}). All three models are able to perform the first task , while only \modz{} and \modzar{} are capable of the second task.

\noindent
\begin{table}[t]
  \vspace*{3ex}
  \caption{\acrshort{1VD} evaluation of \modx{} and \modz/\modzar{} on \textit{VisDial} (v0.9) test set. Results show ranking of answer candidates based on the score functions \(\mathcal{S}_{M}\) and \(\mathcal{S}_{\textit{w2v}}\).}
  \vspace*{-1ex}
  \label{tab:results_ansgen}
  \centering
  \scalebox{0.72}{%
  \begin{tabular}{@{}ccrccccc@{}}
    \toprule
    \makecell{Score\\[-2pt]function} & &  Method & {\bf MR} & {\bf MRR} & {\bf R@1} & {\bf R@5} & {\bf R@10} \\
    \midrule
    \multirow{3}*{\(\mathcal{S}_{M}\)} & & RL-QAbot~\cite{das2017learning} & 21.13 & 0.4370 & - & 53.67 & 60.48 \\
    & & \texttt{MN-QIH-G}~\cite{Das_VisualDialog} & 17.06 & 0.5259 & 42.29 & 62.85 & 68.88 \\
    & & \modx{} ({\sc lw}) & 23.87 & 0.4220 & 30.48 & 53.78 & 57.52 \\
    & & \modx{} ({\sc elbo}) & 20.38 & 0.4549 & 34.08 & 56.18 & 61.11 \\
    \midrule
    \multirow{9}*{\(\mathcal{S}_{\textit{w2v}}\)}
    & & \texttt{MN-QIH-G}~\cite{Das_VisualDialog} & 31.31 & 0.2215 & 16.01 & 22.42 & 34.76 \\
    & & \modx~({\sc recon}) & 15.36 & 0.4952 & 41.77 & 54.67 & 66.90 \\
    & & \modx~({\sc gen}) & {\bf 25.65} & 0.3227 & 25.88 & 33.43 & 47.75 \\
    \cmidrule(l){2-8}
    & & \modz{} & 28.45 & 0.2927 & 23.50 & 29.11 & 42.29 \\
    & \gtQA & \modzar8{} & 25.87 & {\bf 0.3553} & {\bf 29.40} &  {\bf 36.79} & {\bf 51.19}  \\
    & & \modzar10{} & 26.30 & 0.3422 & 28.00 & 35.34 & 50.54 \\
    \cmidrule(l){2-8}
    & & \modz{}  & 30.57 & 0.2188 & 16.06 & 20.88 & 35.37 \\

    & \predA & \modzar8{} & 29.10 & 0.2864 & 22.52 & 29.01 & 48.43 \\

    & &  \modzar10{} & 29.15 & 0.2869 & 22.68 & 28.97 & 46.98  \\
    \bottomrule
  \end{tabular}}
\vspace*{-3ex}
\end{table}

\vspace*{-5ex}
\subsubsection{One-Way Visual Dialogue (\gls{1VD}) task}
\label{sec:vd}
\begin{figure}[t]
  \centering
  \includegraphics[width=0.95\linewidth]{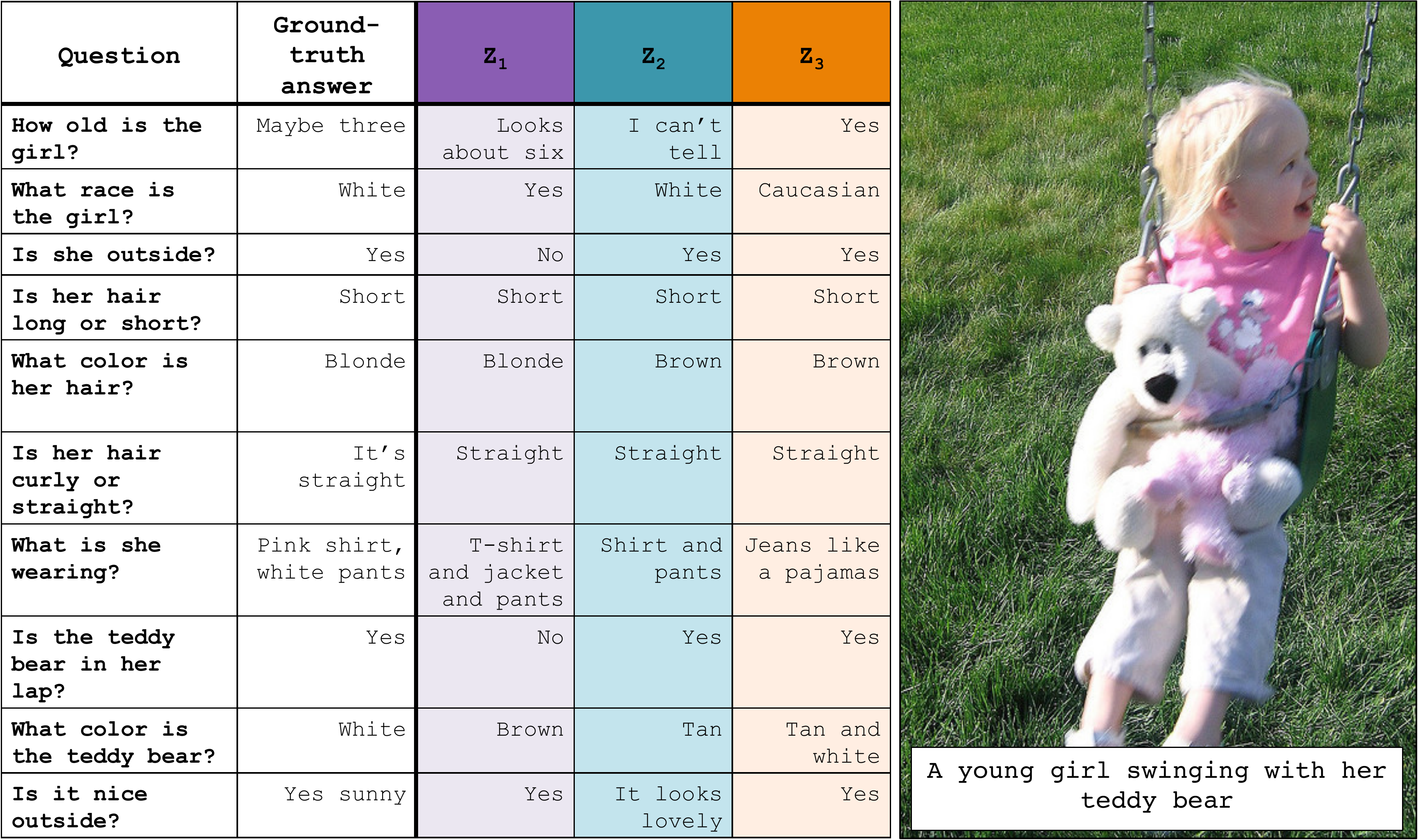}
  \includegraphics[width=0.95\linewidth]{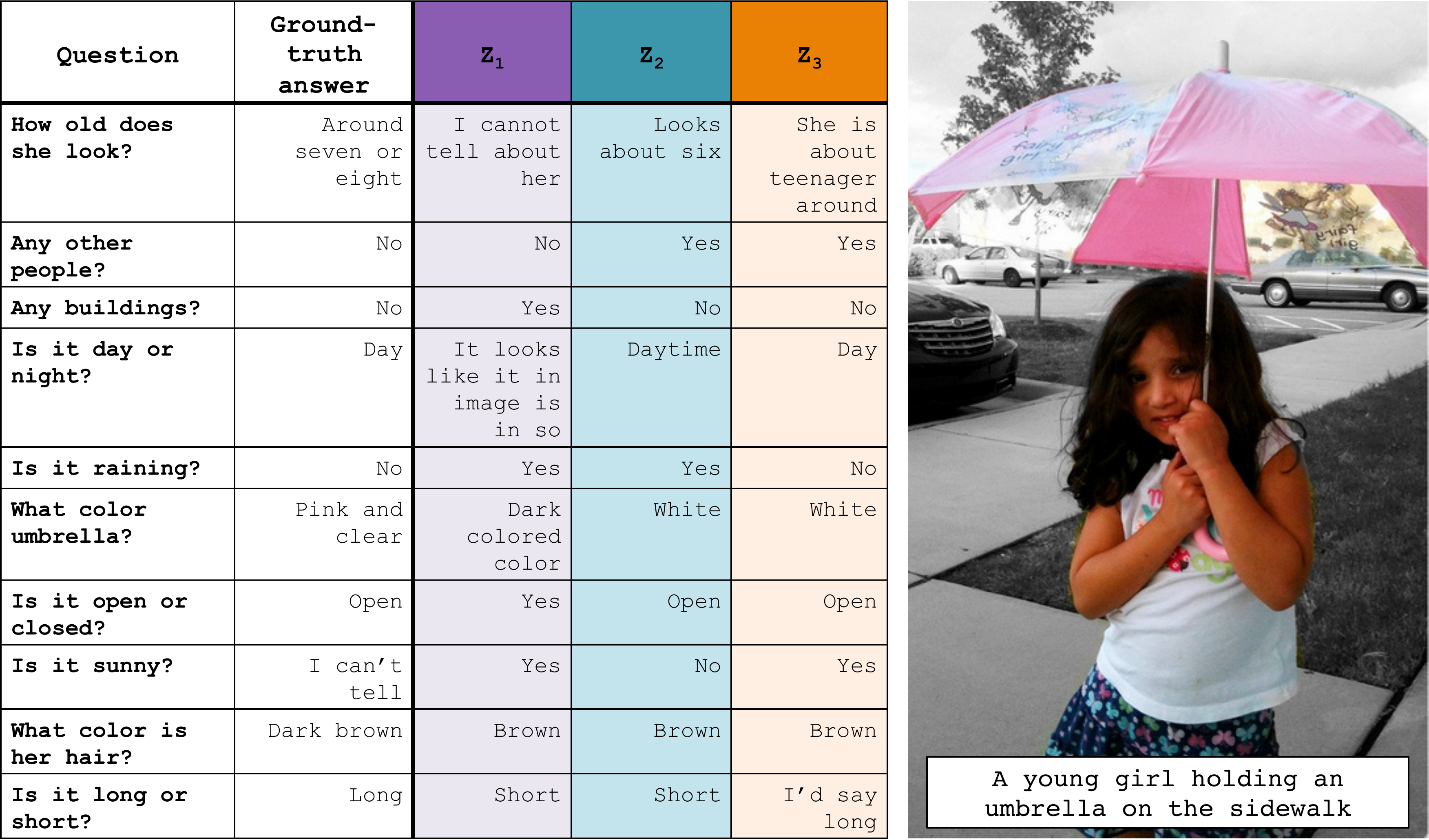}
  \vspace*{-1ex}
  \caption{Example generated answers from \modx's conditional prior -- conditioned on an image, caption, question and dialogue history. See supplement for further examples.}
  \label{fig:qual_gen2}
  \vspace*{-1ex}
\end{figure}

We evaluate the performance of \modx{} and \modz/\modzar{} on \gls{1VD} using the candidate ranking metric of~\cite{Das_VisualDialog} as well as an extension of this which assesses the \textit{generated} answer quality (\cref{tab:results_ansgen}). \cref{fig:gen-vd-hook} and \cref{fig:qual_gen2} show our qualitative results for \gls{1VD}.

\paragraph{Candidate ranking by model log-likelihood [\(\mathbf{\mathcal{S}_{M}}\)]\hfill}
The \textit{VisDial} dataset~\cite{Das_VisualDialog} provides a set of 100 candidate answers \(\{\ans^c_t\}^{100}_{c=1}\) for each question-answer pair at time \(t\) per image.
The set includes the ground-truth answer \(\ans_t\) as well as similar, popular, and random answers.
Das~\etal~\cite{Das_VisualDialog} rank these candidates using the log-likelihood value of each under their model (conditioned on the image, caption and dialogue history, including the current question), and then observe the position of the ground-truth answer (closer to 1 is better). This position is averaged over the dataset to obtain the Mean Rank (MR). In addition, the Mean Reciprocal Rank (MRR; 1/MR) and recall rates at \(k=\{1, 5, 10\}\) are computed.

To compare against their baseline, we rank the 100 candidates answers by estimates of their \emph{marginal} likelihood from \modx{}. This can be done with
\begin{inparaenum}[i)]
\item the conditional \gls{ELBO} (\cref{eq:svqa-g-elbo}), and by
\item likelihood weighting (\textsc{lw}) in the conditional generative model
  \(\p{\ans_t \given \img, \capt, \ctx^{\pl}_t}
  = \int \p{\ans_t, \z \given \img, \capt, \ctx^{\pl}_t} dz
  = \int \p{\z \given \img, \capt, \ctx^{\pl}_t} \p{\ans \given \z, \img, \capt, \ctx^{\pl}_t}\, dz\).
\end{inparaenum}
Ranking by both these approaches is shown in the \(\mathbf{\mathcal{S}_{M}}\) section of \cref{tab:results_ansgen}, indicating that we are comparable to the state of the art in discriminative models of sequential \gls{VQA} \cite{Das_VisualDialog,das2017learning}.

\paragraph{Candidate ranking by \textit{word2vec} cosine distance [\(\mathbf{\mathcal{S}_{\textit{w2v}}}\)]\hfill}
The evaluation protocol of \cite{Das_VisualDialog} scores and ranks a given set of candidate answers, without being a function of the actual answer \emph{predicted} by the model, \(\hat{\ans}_t\).
This results in the rank of the ground-truth answer candidate reflecting its score under the model {\em relative} to the rest of the candidates' scores, rather than capturing the quality of the answer output by the model, which is left unobserved.
To remedy this, we instead score each candidate by the cosine distance between the \textit{word2vec} embedding of the predicted answer \(\hat{\ans}_t\) and that candidate's \textit{word2vec} embedding.
We take the embedding of a sentence to be the average embedding over word tokens following Arora~\etal~\cite{aurora2017simple}.
In addition to accounting for the predicted answer, this method also allows semantic similarities to be captured such that if the predicted answer is similar (in meaning and/or words generated) to the ground-truth candidate answer, then the cosine distance will be small, and hence the ground-truth candidate's rank closer to 1\@.

We report these numbers for \modx{}, iteratively-evaluated \modz/\modzar, and also our baseline model \texttt{MN-QIH-G}~\cite{Das_VisualDialog}, which we re-evaluate using the \textit{word2vec} cosine distance ranking (see \(\mathcal{S}_{\textit{w2v}}\) in \cref{tab:results_ansgen}).
In the case of \modx~({\sc gen}), we evaluate answer \emph{generations} from \modx{} whereby we condition on \(\img, \capt\) and \(\ctx^+_t\) via the prior network, sample \(\z \sim \gaussianp\) and generate an answer via the decoder network. Here we show an improvement of 5.66 points in MR over the baseline.
On the other hand, \modx~({\sc recon}) evaluates answer \emph{reconstructions} in which \(\z\) is sampled from \(\gaussianq\) (where ground-truth answer \(\ans_t\) is provided). We include \modx~({\sc recon}) merely as an ``oracle'' autoencoder, observing its good ranking performance, but do not explicitly compare against it.

We also note that the ranking scores of the block models are worse (by 3-4 MR points) than those of \modx. This is expected since \modx{} is explicitly trained for \gls{1VD} which is not the case for \modz/\modzar.
Despite this, the performance gap between \modx{}~({\sc gen}) and \modz/\modzar{} (with \gtQA) is not large, bolstering our iterative evaluation method for the block architectures.
Note finally that the \modz/\modzar{} models perform better under \gtQA{} than under \predA{} (by 2-3 MR points). This is also expected as answering is easier with access to the ground-truth dialogue history rather than when only the previously \emph{predicted} answers (and ground-truth questions) are provided.

\vspace*{-2.5ex}
\subsubsection{Two-way Visual Dialogue (\gls{2VD}) task}
\label{sec:twowayvd}
\vspace*{-1ex}
Our flexible \gls{CVAE} formulation for visual dialogue allows us to move from \gls{1VD} to the generation of both questions \emph{and} answers (\gls{2VD}).
Despite this being inherently more challenging, \modz/\modzar{} are able to generate diverse sets of questions and answers contextualised by the given image and caption.
\cref{fig:qual_gen1} shows snippets of our two-way dialogue generations.

In evaluating our models for \gls{2VD}, the candidate ranking protocol of~\cite{Das_VisualDialog} which relies on a \emph{given} question to rank the answer candidates, is no longer usable when the questions themselves are being generated.
This is the case for \modz/\modzar{} block evaluation, which has no access to the ground-truth dialogue history, and the \predQA{} iterative evaluation, when the full predicted history of questions and answers is provided (\cref{tab:blockevalmethods}).
We therefore look directly to the \gls{CE} and \gls{KL} terms of the \gls{ELBO} as well as propose two new metrics, \(sim_{\capt,\ques}\) and \(sim_{\circlearrowleft}\), to compare our methods in the \gls{2VD} task:

\begin{figure}[t]
  \begin{center}
    \includegraphics[width=0.95\linewidth]{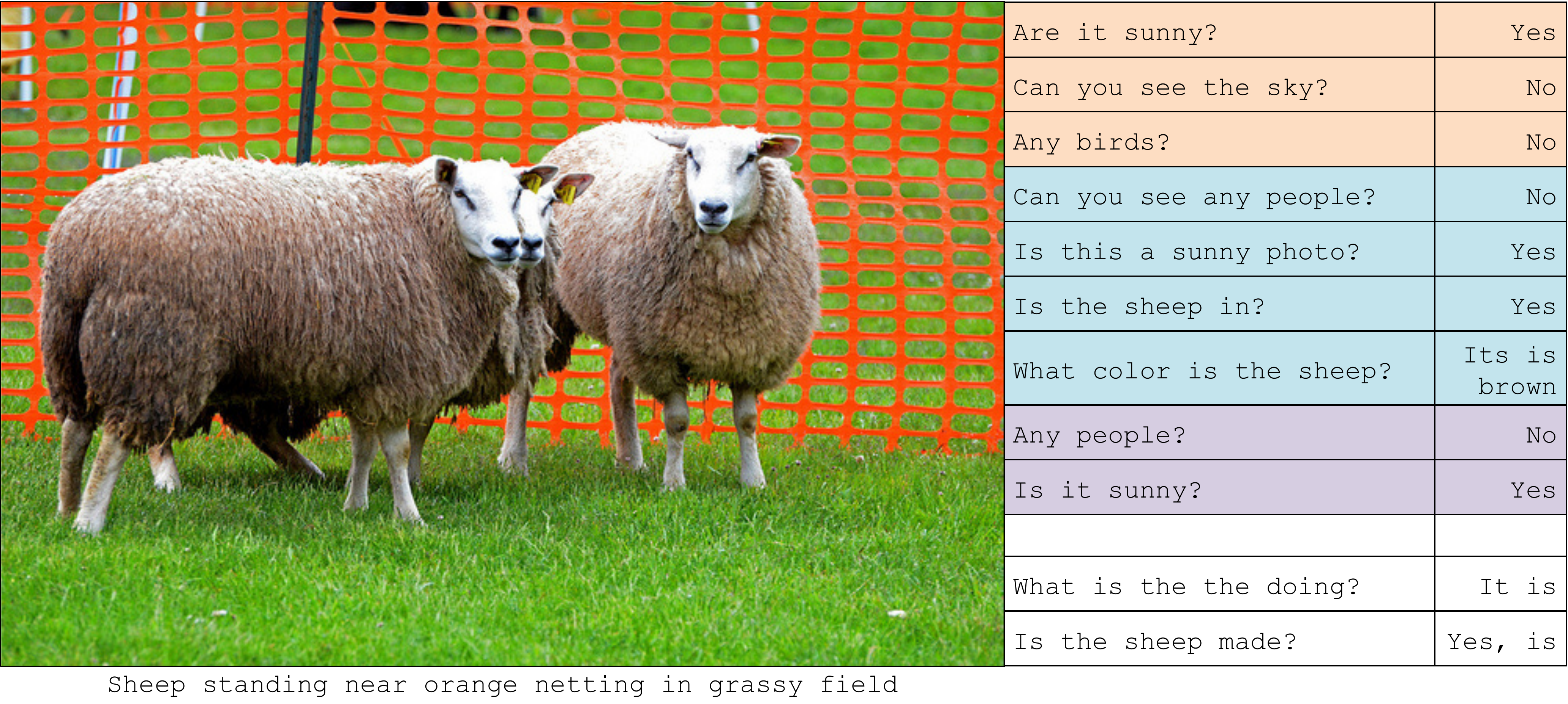}
    \includegraphics[width=0.95\linewidth]{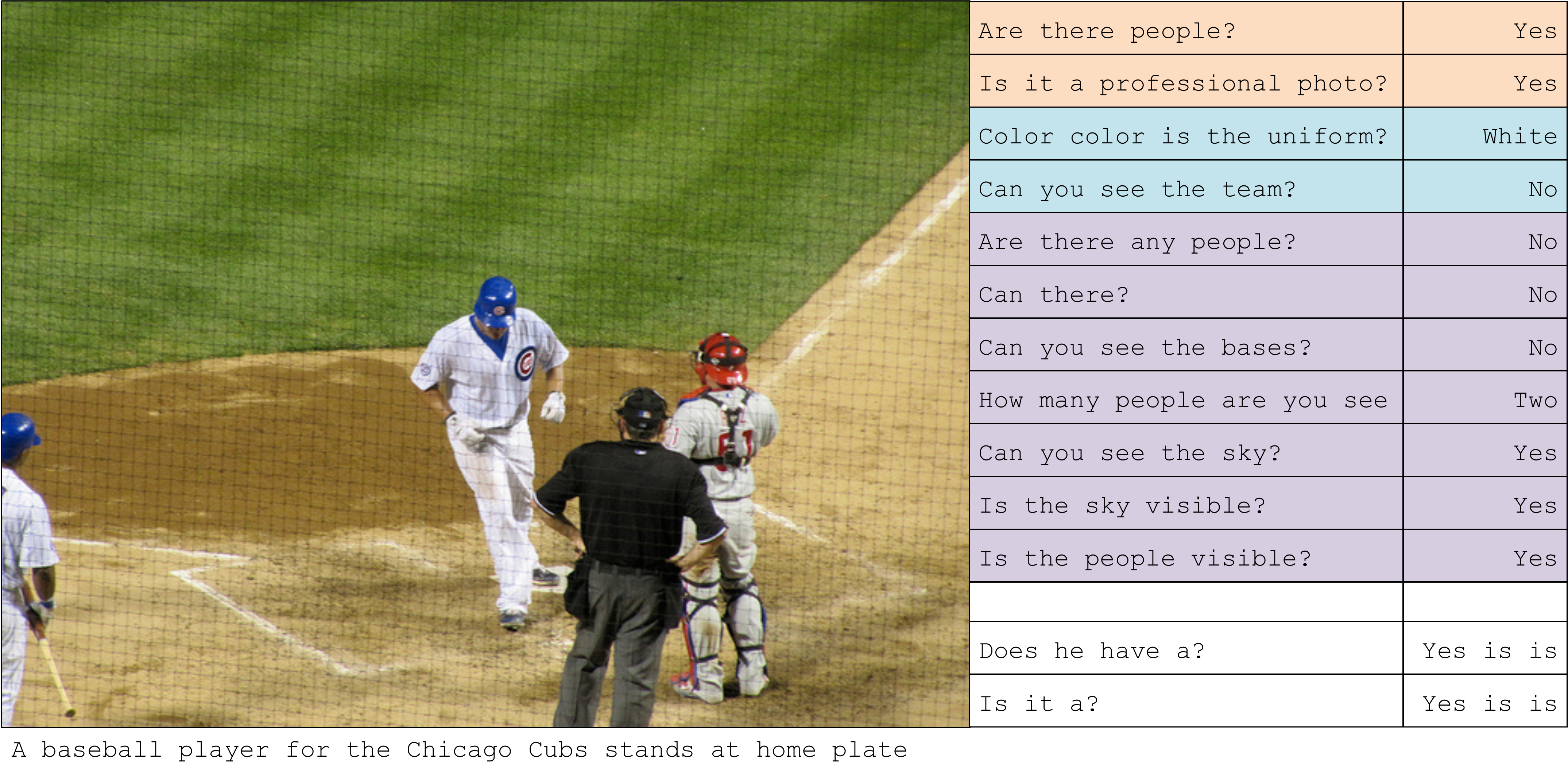}
    \includegraphics[width=0.95\linewidth]{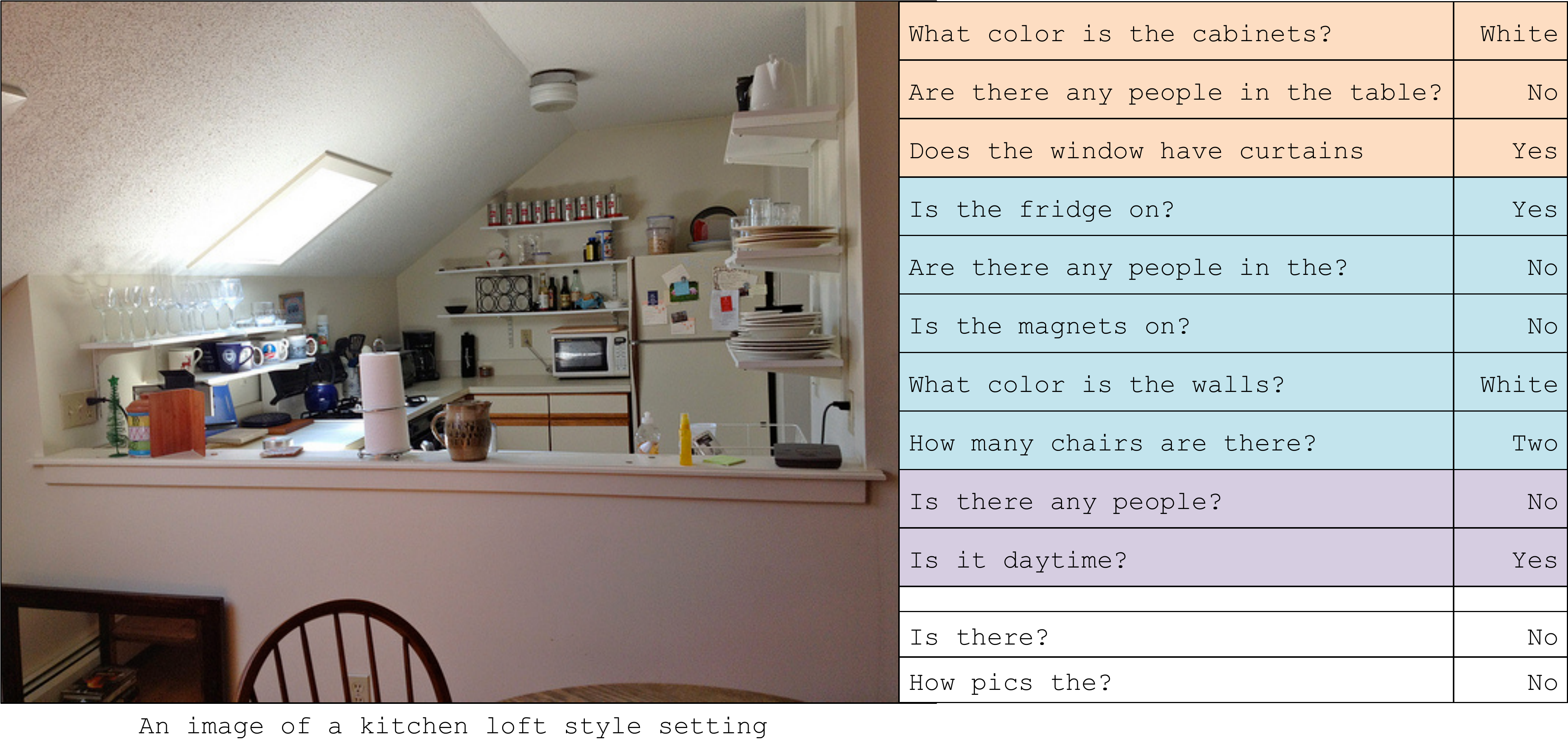}
    \includegraphics[width=0.95\linewidth]{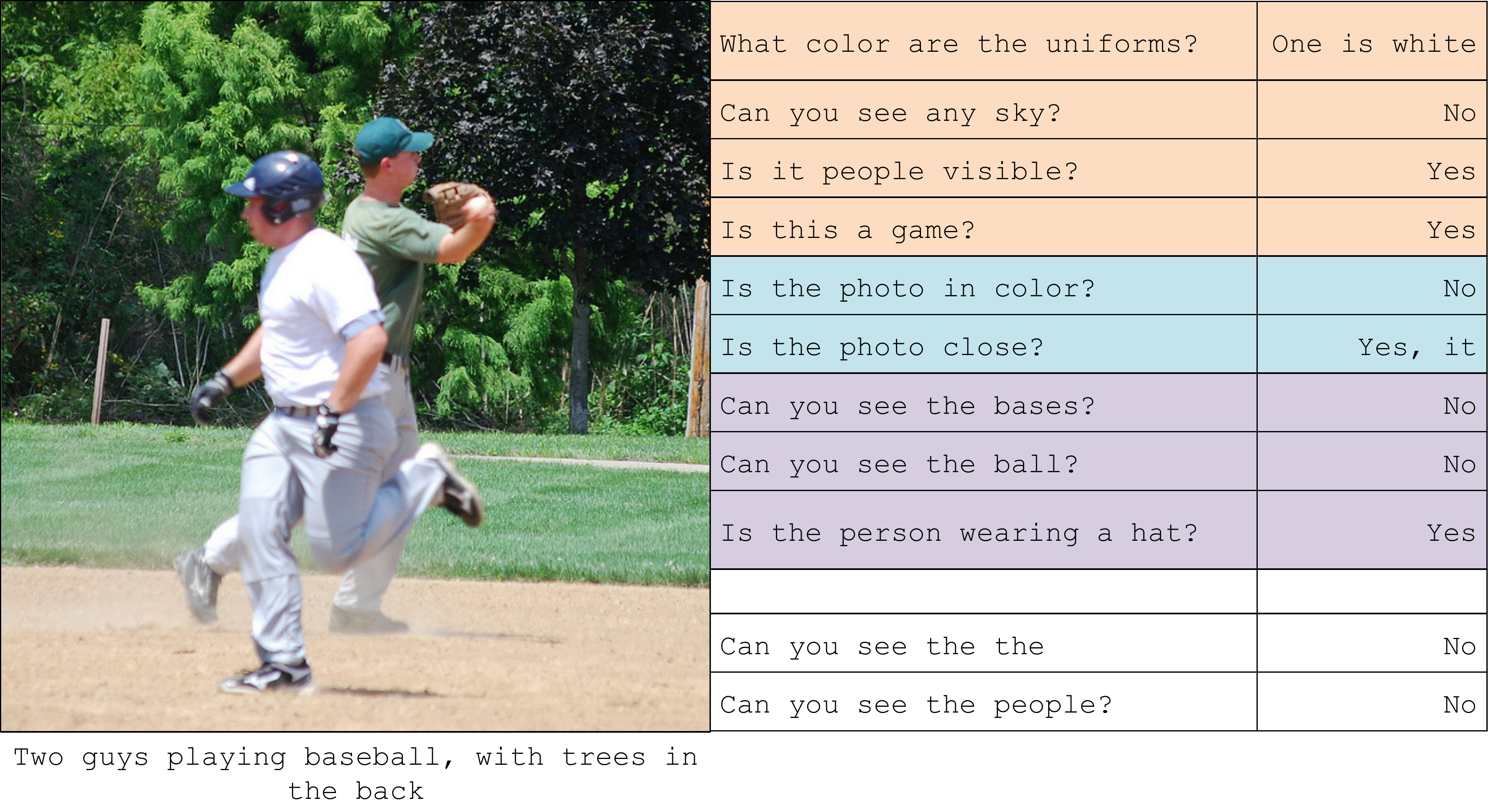}
  \end{center}
  \vspace*{-3.5ex}
  \caption{Examples of two-way dialogue generation from the \modz{}/\modzar{} models. Different colours indicate different generations -- coherent sets with a single colour, and failures in white. See supplement for further examples.}
  \label{fig:qual_gen1}
  \vspace*{-2ex}
\end{figure}

\begin{compactitem}
\item {\bf Question relevance (\(sim_{\capt,\ques}\))}.
We expect a generated question to query an aspect of the image, and we use the presence of semantically similar words in both the question and image caption as a proxy of this. We compute the cosine distance between the (average) \textit{word2vec} embedding of each predicted question \(\ques_t\) and that of the caption \(\capt\), and average over all \(T\) questions in the dialogue (closer to 1 indicates higher semantic similarity).

\item{ \bf Latent dialogue dispersion (\(sim_{\circlearrowleft}\))}.
For a generated dialogue block~\(\dial^g\), \(sim_{\circlearrowleft}\) computes the \gls{KL} divergence \(\KL{q_\phi(\z | \dial^g, \img, \capt)}{q_\phi(\z | \dial, \img, \capt)}\), measuring how close the generated dialogue is to the true dialogue \(\dial\) in the latent space, given the same image~\(\img\) and caption~\(\capt\).
\end{compactitem}

\noindent%
From \cref{tab:results_modz}, we observe a decrease in the loss terms as the auto-regressive capacity of the model increases (none \to{} 8 \to{} 10), suggesting that explicitly enforcing sequentiality in the dialogue generations is useful.
For \(\text{sim}_{\circlearrowleft}\) within a particular model, the dispersion values are typically larger for the harder task (without dialogue context).
We also observe that dispersion increases with number of \textsc{AR} layers, suggesting {\sc AR} improves the diversity of the model outputs, and avoids simply recovering data observed at train time.

\begin{table}[t]
  \vspace*{3ex}
  \caption{\acrshort{2VD} evaluation on \textit{VisDial} (v0.9) test set for \modz{}/\modzar{} models. For \(\dial\), `\(\emptyset\)' indicates block evaluation, and `\predQA' indicates iterative evaluation (see \Cref{sec:blockevalmethods}).}
  \label{tab:results_modz}
  \centering
  \vspace*{-1ex}
  \scalebox{0.9}{%
  \begin{tabular}{@{}rccccr@{}}
    \toprule
    Method                   & \(\dial\) & {\bf CE} & {\bf KLD} & $\text{sim}_{\capt,q}$ & $\text{sim}_{\circlearrowleft}$ \\
    \midrule
    \multirow{2}*{\modz}     & \(\emptyset\)   & 31.18 & 4.34 & 0.4931 & 14.20 \\
                             & \predQA & 25.40 & 4.01 & 0.4091 & 1.86  \\
    \midrule
    \multirow{2}*{\modzar8}  & \(\emptyset\)   & 28.81 & 2.54 & 0.4878 & 31.50 \\
                             & \predQA & 26.60 & 2.29 & 0.3884 & 2.39  \\
    \midrule
    \multirow{2}*{\modzar10} & \(\emptyset\)   & 28.49 & 1.89 & 0.4927 & 44.34 \\
                             & \predQA & 24.93 & 1.80 & 0.4101 & 2.35  \\
    \bottomrule
  \end{tabular}}
  \vspace*{-3ex}
\end{table}

While the proposed metrics provide a novel means to evaluate dialogue in a generative framework, like all language-based metrics, they are not complete.
The question-relevance metric, \(\text{sim}_{\capt,\ques}\), can stagnate, and neither metric precludes redundant or nonsensical questions.
We intend for these metrics to \emph{augment} the bank of metrics available to evaluate dialogue and language models.
Further evaluation, including
\begin{inparaenum}[i)]
\item using auxiliary tasks, as in the image-retrieval task of~\cite{das2017learning}, to drive and evaluate the dialogues, and
\item turning to human evaluators to rate the generated dialogues,
\end{inparaenum}
can be instructive in painting a more complete picture of our models.

\vspace*{-1.5ex}
\section{Conclusion}
\vspace*{-1.5ex}
In this work we propose \fd{}, a generative convolutional model for visual dialogue which is able to generate answers (\gls{1VD}) as well as generate both questions \emph{and} answers (\gls{2VD}) based on a visual context. In the \gls{1VD} task, we set new state-of-the-art results with the answers generated by our model, and in the \gls{2VD} task, we are the first to establish a baseline, proposing two novel metrics to assess the quality of the generated dialogues. In addition, we propose and evaluate our models under a much more realistic setting for both visual dialogue tasks in which the \emph{predicted} rather than ground-truth dialogue history is provided at test time. This challenging setting is more akin to real-world situations in which dialogue agents must be able to evolve with their predicted exchanges. We emphasize that research focus must be directed here in the future. Finally, under all cases, the sets of questions and answers generated by our models are qualitatively good: diverse and plausible given the visual context.
Looking forward, we are interested in exploring additional methods for enforcing diversity in the generated questions and answers, as well as extending this work to explore \emph{recursive} models of reasoning for visual dialogue.

\paragraph{Acknowledgements}
This work was supported by the EPSRC, ERC grant ERC-2012-AdG 321162-HELIOS, EPSRC grant Seebibyte EP/M013774/1, EPSRC/MURI grant EP/N019474/1 and the Skye Foundation.

{\small
\bibliographystyle{ieee}
\bibliography{references}
}

\newpage
\appendix

\glsreset{1VD}
\glsreset{2VD}
\section{Glossary}
\label{sec:glossary}

\begin{description}
\item[block dialogue/architecture] Models \modz/\modzar{} are built and trained for the task of \gls{2VD} with data~\(\x = \dial\) and condition variable~\(\C = \grpB{\img, \capt}\). Since \(\dial\) refers to the whole dialogue sequence/block \(\grpA{(\ques_t, \ans_t)}^T_{t=1}\) we refer to \modz/\modzar{} as {\em block} architectures.
\item[generation]  This represents the scenario when only the condition variable \(\C\) is available at test time. In this case, the decoder network receives a sample \(\z \sim \p{\z \given \C}\), a multivariate Gaussian parametrised by \(\mup\) and exponentiated \(\logvarp\) learned using the {\em prior} network. We call the decoded output~\(\hat{\dial}\) a {\em generation}. 
\item[reconstruction] Differing from a generation, both \(\C\) {\em and} \( \dial \) are available. The decoder network receives a sample \(\z \sim \q{\z \given \dial, \C}\), a multivariate Gaussian parametrised by \(\muq\) and exponentiated \(\logvarq\) learned using the {\em encoder} network. We call the decoded output \(\hat{\dial}\) a {\em reconstruction}. The reconstruction pipeline is used during training when the input \(\dial\) {\em and} the condition variable~\(\C\) are available. Note, this pipeline is also used when \modz/\modzar{} are evaluated iteratively (see \cref{sec:blockevalmethods}).
\end{description}

\section{Extended Quantitative Results on \acrshort{1VD} task}
\label{sec:vd_supp}

\Cref{tab:results_ansgen} in the main paper evaluates \modx{} and \modz/\modzar{} in the task of \gls{1VD}. Here we shed light on these numbers and the metrics used to obtain them. We also present a more extensive quantitative analysis of \modz/\modzar{} in the \gls{1VD} task (see \cref{tab:results_modz_predA_supp}).
\paragraph{Evaluating \modz/\modzar{} on \gls{1VD}}
We extend \cref{tab:results_ansgen} with\cref{tab:results_modz_predA_supp}, which further compares \modz/\modzar{} under the iterative evaluation settings of \gtQA{} and \predA, using the \gls{CE} and \gls{KL} terms of the \gls{ELBO} and our two new metrics, \(\text{sim}_{\capt,\ques}\) and  \(\text{sim}_{\circlearrowleft}\).
We observe that \modz/\modzar{} (\gtQA) shows superior performance of around 7-10 points in MR over \modz/\modzar{} (\predA), and also improves in MRR and recall rates.
This is expected since the ground-truth rather than predicted answers are included in the dialogue history (along with the ground-truth questions).
The metrics \(\text{sim}_{cap,q}\) and \(\text{sim}_{\circlearrowleft}\), on the other hand, show very little performance difference across the two evaluation settings.
We also note that ranking performance is worse when both image \(\img\) and caption \(\capt\) are excluded from condition variable.
This does not, however, correlate with the \gls{CE} and \gls{KL} terms of the loss which are lower for a condition-less setting. We attribute this to the model being transformed from a \acrshort{CVAE} to a \acrshort{VAE}, hence lifting the burden of capturing the conditional posterior distribution (i.e. the \gls{KL} is now between an unconditional \(\q{\z \given \x}\) and \(\mathcal{N}(0,1)\)).
Interestingly, however, excluding either the image or the caption achieves similar performance to when both are included, indicating that the caption acts as a good textual proxy of the image (a reassurance of our \(\text{sim}_{\capt,\ques}\) metric).
\begin{table*}[h]
  \caption{\acrshort{1VD} evaluation of \modz/\modzar{} on \textit{VisDial} (v0.9) test set. Results show ranking of answer candidates based on the \(\mathcal{S}_{\textit{w2v}}\) scoring function. Note that \gtQA{} indicates the iterative evaluation method when the {\em ground-truth} dialogue history is provided, while \predA{}, the iterative evaluation method when the ground-truth question and \emph{predicted} answer history is provided (see \Cref{sec:blockevalmethods}). The \pl{} and \mn{} indicate models \textit{trained} with and without respective conditions, image \(\img\) and caption \(\capt\).}
  \label{tab:results_modz_predA_supp}
  \centering
  \scalebox{1.0}{%
  \begin{tabular}{@{}rcccccccccccr@{}}
    \toprule
    Method  & \(\img\) & \(\capt\) & & {\bf CE} & {\bf KLD} & {\bf MR} & {\bf MRR} & {\bf R@1} & {\bf R@5} & {\bf R@10} & $\text{sim}_{cap,q}$ & $\text{sim}_{\circlearrowleft}$ \\
    \midrule
    \multirow{5}*{ \modz} & \pl & \pl & \gtQA  & 18.87 & 4.36  & 28.45 & 0.2927 & 23.50 & 29.11 & 42.29 & 0.4374 &  2.68 \\
    \cmidrule{4-13}
    						 & \pl & \pl & \multirow{4}*{\predA} & 25.10 & 4.02 & 30.57 & 0.2188 & 16.06 & 20.88 & 35.37 & 0.4118 & 2.42  \\
    						 & \mn & \pl &  &     						 16.80  & 3.13 & 27.76 & 0.3243 & 26.59 & 33.21 & 47.65 & 0.4491 & 4.48 \\
    						 & \pl & \mn & & 21.02 & 4.71 & 29.82 & 0.2144 & 15.25 & 21.07 & 34.96 & 0.4551 & 5.44 \\
    						 & \mn & \mn &  & 19.35 & 13.34 & 29.00 & 0.3026 &  24.36 & 30.70 & 47.62  & 0.4638 & 6.17 \\
    	\midrule

    	\multirow{5}*{ \modzar8} & \pl & \pl & \gtQA &  15.11 &  2.53 & 25.87 & 0.3553 & 29.40 & 36.79 & 51.19 & 0.4703 & 4.30 \\
    	\cmidrule{4-13}

    						 & \pl & \pl & \multirow{4}*{\predA} & 25.70 & 2.21 & 29.10 & 0.2864 & 22.52 & 29.01 & 48.43 & 0.3885 & 3.47  \\
    						 & \mn & \pl &  & 16.19 & 2.80  & 26.04 & 0.3566 & 29.62 & 36.75 & 50.62 & 0.4626 & 4.17 \\
    						 & \pl & \mn &  & 20.39 & 2.89 & 28.99 & 0.3024 & 24.33 & 30.74 & 47.17 &  0.4461 & 8.16  \\
    						 & \mn & \mn &  & 20.92 & 2.84 & 28.79 & 0.3045 & 24.46 & 30.99 & 48.10 & 0.4442 & 0.18 \\
    \midrule
    	\multirow{5}*{ \modzar10} & \pl & \pl & \gtQA & 16.04& 1.89 & 26.30  & 0.3422& 28.00 & 35.34 & 50.54  & 0.4708 & 4.84  \\
    \cmidrule{4-13}
    					      & \pl & \pl & \multirow{4}*{\predA} & 24.77 & 1.81 & 29.15 & 0.2869 & 22.68 & 28.97 & 46.98 & 0.4058 & 2.85 \\
    						 & \mn & \pl & & 19.97 & 2.58 & 26.84 & 0.3212 & 25.90 & 32.92 & 47.68  & 0.4424 & 5.95 \\
    						 & \pl & \mn &  & 20.39 & 2.79 & 27.27 & 0.3157 & 25.45 & 32.26 & 47.87 & 0.4707 & 13.22  \\
    						 & \mn & \mn & & 19.17 & 0.00 & 29.00 & 0.3026 & 24.36 & 30.70 & 47.62 & 0.4614 & 0.00\\
    \bottomrule
  \end{tabular}}
\end{table*}

\section{Extended Quantitative Results on \acrshort{2VD} task}

Extending \cref{tab:results_modz} in the main paper, \cref{tab:results_modz_supp} here shows results for \modz/\modzar{} trained with permutations of the image \(\img\) and caption \(\capt\)) (denoted by \pl{} if included in the condition, and \mn{} otherwise).
We note the decrease in \gls{CE} and \gls{KL} as conditions (\(\img, \capt\)) are excluded from the model. This is expected since the task of dialogue generation is made simpler without the constrains of an explicit visual/textual condition.
\begin{table}[h]
  \vspace*{3ex}
  \caption{\acrshort{2VD} evaluation on \textit{VisDial} (v0.9) test set for \modz{}/\modzar{} models. Note that \(\dial\) left blank indicates the block evaluation method, when a whole dialogue is generated given only an image and its caption, while \predQA{} indicates the iterative evaluation method when previously generated questions \emph{and} answers are included in the dialogue history (see Section~\ref{sec:blockevalmethods}). The \pl{} and \mn{} indicate models \textit{trained} with and without respective conditions, image \(\img\) and caption \(\capt\).}
  \label{tab:results_modz_supp}
  \centering
  \vspace*{-1ex}
  \scalebox{0.9}{%
  \begin{tabular}{@{}cccccccr@{}}
    \toprule
    Method  & \(\img\) & \(\capt\) & \(\dial\) & {\bf CE} & {\bf KLD} & $\text{sim}_{\capt,q}$ & $\text{sim}_{\circlearrowleft}$ \\
    \midrule
    \multirow{8}*{\modz} & \multirow{2}*{\pl} & \multirow{2}*{\pl} & \(\emptyset\) & 31.18 & 4.34 & 0.4931 & 14.20  \\
            &  & &  \predQA & 25.40 & 4.01 & 0.4091 & 1.86\\
    \cmidrule(l){2-8}
            & \multirow{2}*{\mn} & \multirow{2}*{\pl} & \(\emptyset\) & 29.09 & 3.26 & 0.4889 & 11.23 \\
            &  & &  \predQA & 24.59 & 3.05 & 0.3877 & 3.45\\
    \cmidrule(l){2-8}
            & \multirow{2}*{\pl} & \multirow{2}*{\mn} & \(\emptyset\) & 28.60 & 4.26 & 0.4634 & 15.56  \\
            &  & &  \predQA & 29.85 & 4.66 & 0.4221 & 3.54\\
    \cmidrule(l){2-8}
            & \multirow{2}*{\mn} & \multirow{2}*{\mn} & \(\emptyset\) & 19.92 & 6.42 & 0.4590 & 6.34 \\
            &  & &  \predQA & 19.34 & 0.00 & 0.4638 & 0.00\\
    \midrule
    \multirow{8}*{\modzar8} & \multirow{2}*{\pl} & \multirow{2}*{\pl} & \(\emptyset\) & 28.81 & 2.54 & 0.4878 & 31.50 \\
            &  & &  \predQA & 26.60 & 2.29 & 0.3884 & 2.39 \\
    \cmidrule(l){2-8}
            & \multirow{2}*{\mn} & \multirow{2}*{\pl} & \(\emptyset\) & 30.59 & 2.72 & 0.4889 & 43.17  \\
            &  & &  \predQA & 26.15 &  2.77 & 0.3758 & 3.57  \\
    \cmidrule(l){2-8}
            & \multirow{2}*{\pl} & \multirow{2}*{\mn} & \(\emptyset\) & 31.51 & 2.91 & 0.4602 & 24.75   \\
            &  & &  \predQA & 21.41 &  2.68 & 0.4453 & 5.49 \\
    \cmidrule(l){2-8}
            & \multirow{2}*{\mn} & \multirow{2}*{\mn} & \(\emptyset\) &  20.32 & 2.77 & 0.4464 & 0.26 \\
            &  & &  \predQA & 21.53 & 2.99 & 0.4419 & 0.10 \\
    \midrule
    \multirow{8}*{\modzar10} & \multirow{2}*{\pl} & \multirow{2}*{\pl} & \(\emptyset\) &  28.49 & 1.89 & 0.4927 & 44.34 \\
            &  & & \predQA & 24.93 & 1.80 & 0.4101 & 2.35\\
    \cmidrule(l){2-8}
            & \multirow{2}*{\mn} & \multirow{2}*{\pl} & \(\emptyset\) & 30.83 & 2.53 & 0.4951 & 38.60 \\
            &  & &  \predQA & 28.59 & 2.52 & 0.3903 & 1.91\\
    \cmidrule(l){2-8}
            & \multirow{2}*{\pl} & \multirow{2}*{\mn} & \(\emptyset\) & 30.18 & 2.89 & 0.4592 & 100.81\\
            &  & &  \predQA &  28.32 & 2.44 & 0.4334 & 6.73\\
    \cmidrule(l){2-8}
            & \multirow{2}*{\mn} & \multirow{2}*{\mn} & \(\emptyset\) & 19.60 & 0.00 & 0.4585 & 0.00  \\
            &  & &  \predQA & 19.17 & 0.00 & 0.4614 & 0.00\\
    \bottomrule
  \end{tabular}}
  \vspace*{-3ex}
\end{table}

\section{Network architectures and training}

The following section provides detailed descriptions of the architectures of our models \modx, \modz{} and \modzar. The descriptions are dense but thorough. We also include further details of our training procedure. Where not explicitly noted, each convolutional layer is proceeded by a batch normalisation layer (with momentum \(=0.001\) and learnable parameters) and a \textit{ReLU} activation.

\paragraph{Prior network} The prior neural network, parametrised by \(\theta\), takes as input the image \(\img\), the caption \(\capt\) and the dialogue context. For the model \modx, this context is \(\ctx^{\pl}_t\), containing the dialogue history up to \(t\text{-}1\) and the current question \(\ques_t\). For models \modz/\modzar, the dialogue context is the null set (\(\ctx=\emptyset\)).
To obtain the image representation, we scale and centre-cropped each image to \(3 \times 224 \times 224\) and feed it through \textit{VGG-16}~\cite{Simonyan_2014c}. The output of the penultimate layer is extracted and \(\ell_2\)-normalised (as in \cite{Das_VisualDialog}) to obtain a $4096$-dimensional image feature vector.
For the caption, we pass \(\capt\) through a pre-trained \textit{word2vec}~\cite{mikolov2013distributed} model (we do not learn these word embeddings) to obtain \(\e{\capt} \in \mathbb{R}^{300 \times L}\) where \(L\) is the maximum sentence length (\(L=64\)).
For the dialogue context (relevant only in the case of \modx{}) we pass the one-hot encoding of each word through a learnable word embedding module. We stack these embeddings as described in \cref{sec:vd-conv} of the main paper to obtain \(\e{\ctx^\pl_t} \in \mathbb{R}^{E \times L \times K}\), where \(E\) is the word embedding dimension (\(E=256\)), \(L\) is the maximum sentence length (\(L=64\)) and \(K\) is the number of dialogue entries at time \(t\).
We encode these inputs convolutionally to obtain \(\C\) (the encoded condition) as follows: \(\e{\capt}\) is passed through a convolutional block (output size \(64\times 8\times 8\)) and concatenated with the image feature vector (reshaped to \(64\times 8\times 8\)). The concatenated output is passed through a convolutional block to obtain the jointly encoded image-caption (output size \(64\times 8\times 8\)). If \(\ctx \neq \emptyset\), then the context is passed through a convolutional block (output size \(64\times 8\times 8\)) and is concatenated with the encoded image-caption and passed through yet another convolutional block to get the encoded image-caption-context (output size \(64\times 8\times 8\)). We call this the encoded condition \(\C\).
The encoded condition \(\C\) is then passed through a further convolutional block (output size \(256\times 4\times 4\)) followed by two final convolutional layers (in parallel) to obtain \(\mup\) and \(\logvarp\), respectively, the parameters of the conditional prior \(\p{\z \given \C}\). At this stage, \(\mup\) and \(\logvarp\) are both of size \(512\times 1\times 1\) (the latent dimensionality).
%
%
At test time, a sample is obtained via \(\z \sim \gaussianp\) and is passed to the decoder in order to generate a sample \(\hat{\ans_t}\) (for \modx) or \(\hat{\dial}\) (for \modz/\modzar{}).

\paragraph{Encoder network}
The encoder network, parametrised by \(\phi\), takes \(\x\) as input along with the encoded condition, \(\C\), obtained from the prior network.
For model \modx, \(\x=\ans_t\) and \(\C=\{\img, \capt, \ctx^{\pl}_t\}\). For models \modz/\modzar, \(\x = \dial = \grpA{(\ques_t, \ans_t)}^T_{t=1}\) and \(\C=\{\img, \capt\}\).
In all models, \(\x\) is passed through a learnable word embedding module, and the word embeddings stacked (see \cref{sec:vd-conv} in the main paper) to obtain \(\e{\x} \in \mathbb{R}^{E \times L \times M}\), where \(E=256\), \(L=64\) and \(M\) is the number of entries in \(\x\) (for \modx{}, \(M=1\) and for \modz/\modzar{} \(M=2T\)). In this way, we transform \(\x\) into a single-channel answer `image' in the case of \modx, and a multi-channel image of alternating questions and answers in the case of \modz/\modzar. \(\e{\x}\) is then passed through a convolutional block (output size \(64\times 8\times 8\)), the output of which is concatenated with \(\C\) and forwarded through another convolutional block (output size \(256\times 4\times 4\)). This output is forwarded through two final convolutional layers (in parallel) to obtain \(\muq\)  and \(\logvarq\), the parameters of the conditional latent posterior \(\q{\z \given \x, \C}\). Here \(\muq\) and \(\logvarq\) are both of size \(512\times 1\times 1\).

At train time, the \gls{KL} divergence term of the \gls{ELBO} is computed using \(\{ \muq,\stddevq \}\) (from the encoder network) and \(\{ \mup,\stddevp \}\) (from the prior network).

\paragraph{Decoder network}
The decoder network (for simplicity, the parameters of the prior and decoder network are subsumed into \(\theta\)) takes as input a latent \(\z\) and the encoded condition \(\C\). During training, \(\z\) is sampled from a Gaussian parametrised by the \(\muq\) and exponentiated \(\logvarq\) outputs of the {\em encoder} network. This distribution is \(\q{\z \given \x, \C}\). At test time, \(\z\) is sampled from a Gaussian parametrised by the \(\mup\) and exponentiated \(\logvarp\) outputs of the {\em prior} network. This distribution is \(\p{\z \given \C}\).
At both train and test time, we employ the commonly-used `re-parametrisation trick'~\cite{Kingma_2014auto} to compute the latent sample as \(\z = \mathbb{\mu} + \epsilon\mathbb{\sigma}\) where \(\epsilon \sim \mathcal{N}(0,1)\) and \(\mu\) and \(\sigma\) correspond to those derived from the encoder or prior network as described above.

The sample \(\z\) is then transformed through a transpose-convolutional block (output size \(64 \times 8 \times 8\)), concatenated with \(\C\) and forwarded through a convolutional block (output size \(64 \times 8 \times 8\)). This output is forwarded through a second transpose-convolutional block, producing an intermediate output volume of dimension \(M \times E \times L\) which we permute to match the size of \(\e{\x}\). As before, \(E=256\), \(L=64\) and \(M=1\) (for \modx) or \(M=2T\) (for \modz/\modzar).

Following this, our models diverge in architecture: \modx{} and \modz{} employ a standard linear layer which projects the \(E\) dimension of the intermediate output to the vocabulary size \(V\). The \modzar{} model instead employs an autoregressive module (detailed below) followed by this standard linear layer. At train time, the \(V\)-dimensional network output is \textit{softmax}ed and used in the computation of the \gls{CE} term of the \gls{ELBO}. At test time, the \(\mathop{argmax}\) of the (\textit{softmax}-ed) output is taken to be the index of the word token predicted.
We share the weight matrices of the decoder's final linear layer and the encoder and prior's learnable word embedding module  (which are the same size by virtue of our network architecture) with the motivation that language encoders and decoders should share common word representations.

\paragraph{Autoregressive block}
The autoregressive ({\sc AR}) block (\(AR\text{-}N\) in \cref{fig:nw-arch} - bottom) in \modzar's decoder is inspired by \textit{PixelCNN}~\cite{oord2016pixelrnn} which sequentially predicts the pixels in an image along the two spatial dimensions. In the same fashion, we use an autoregressive approach to sequentially predict the next sentence (question or answer) in a dialogue. Since our framework is convolutional with sentences viewable as `images', our approach can similarly be adapted from that of~\cite{oord2016pixelrnn, gulrajani2016pixelvae}. We first reshape the intermediate output of the decoder to \(E \times L\!*\!M\) (essentially `unravelling' the dialogue sequentially into a stack of its word embeddings). We then apply a size-preserving masked convolution to the reshaped output (followed by a learnable batch normalisation and a \textit{ReLU} activation). We call this triplet an AR layer. The masked convolution of the {\sc AR} layer ensures that future rows (i.e. future \(E\)-dimensional word embedding) are hidden in the prediction of the current row/word embedding. We apply \(N\) {\sc AR} layers in this way with each layer taking in the output of the previous {\sc AR} layer. Following the \(AR\text{-}N\) block, a linear layer projects the final output's \(E\) dimension to the vocabulary size \(V\). We report numbers for \(N=\{8, 10\}\). We base our implementation of the {\sc AR} block on a publicly-available implementation of \textit{PixelCNN}.

\vspace*{-1ex}
\section{Dialogue preprocessing}
\vspace*{-1ex}
The word vocabulary is constructed from the \textit{VisDial} v0.9~\cite{Das_VisualDialog} training dialogues (not including the candidate answers). The dialogues are preprocessed as follows: apostrophes are removed, numbers are converted to their worded equivalents, and all exchanges are made lower-case and either padded or truncated to a maximum sequence length (\(L=64\)). The vocabulary is also filtered such that words with a frequency of \(<\)5 are removed and replaced with the {\small\texttt{UNK}} token. After pre-processing and filtering, the vocabulary size is \(V = 9710\).

\vspace*{-1ex}
\section{Extended Qualitative Results}
\begin{figure*}[!h]
  \centering
  \begin{tabular}{@{}c@{}}
    \includegraphics[width=0.884\linewidth]{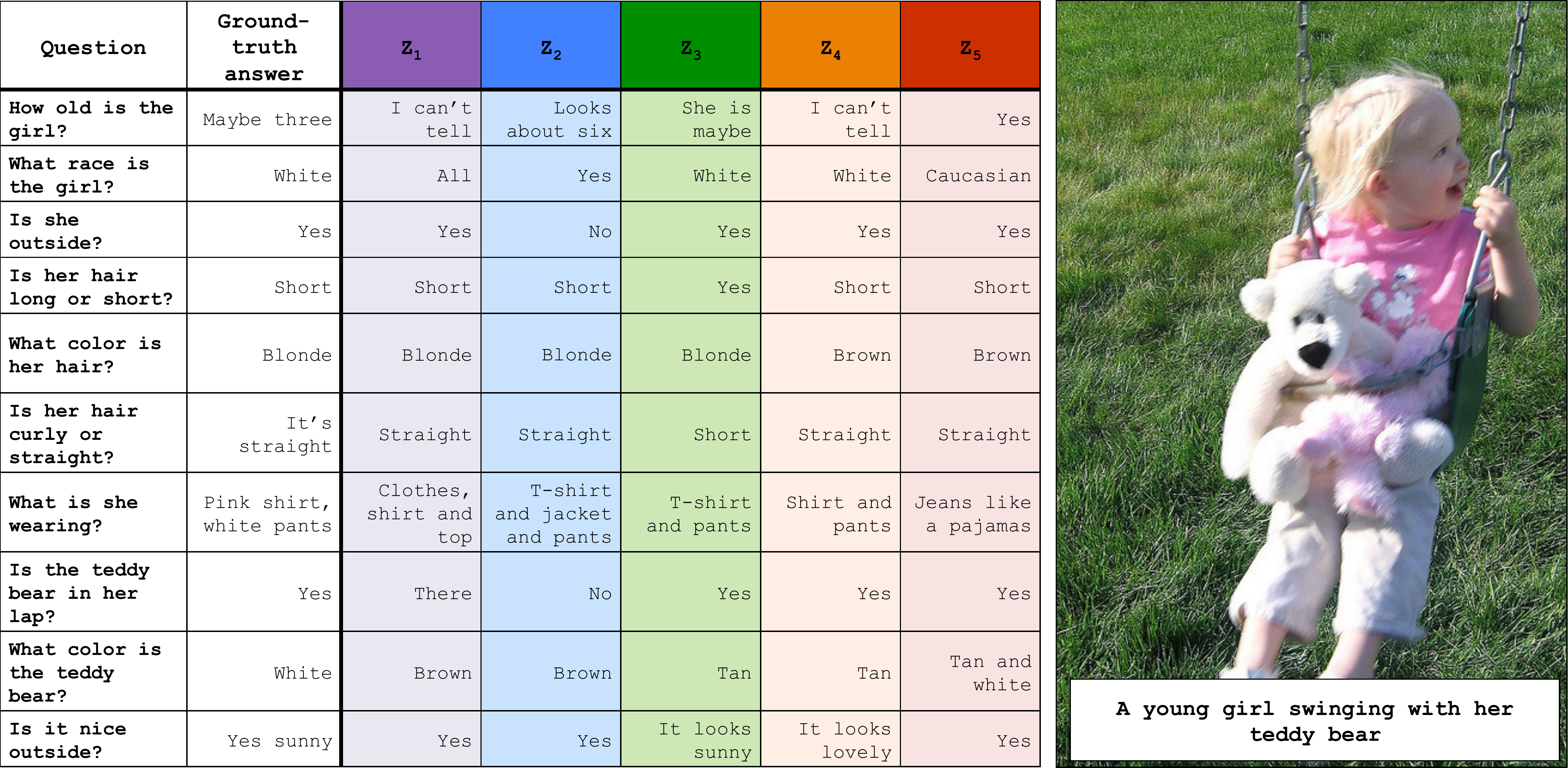}\\
    \includegraphics[width=0.884\linewidth]{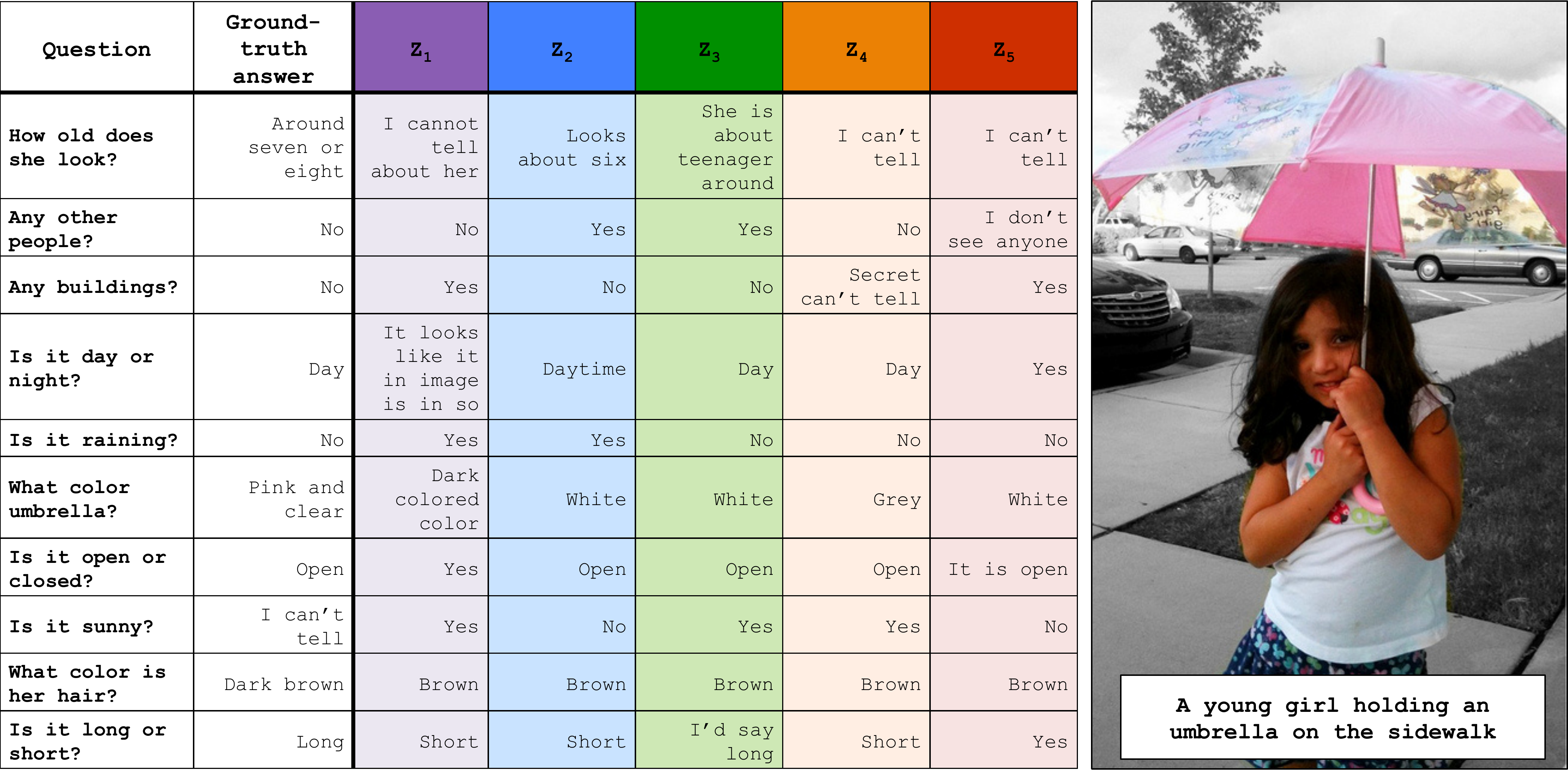}\\
    \includegraphics[width=0.884\linewidth]{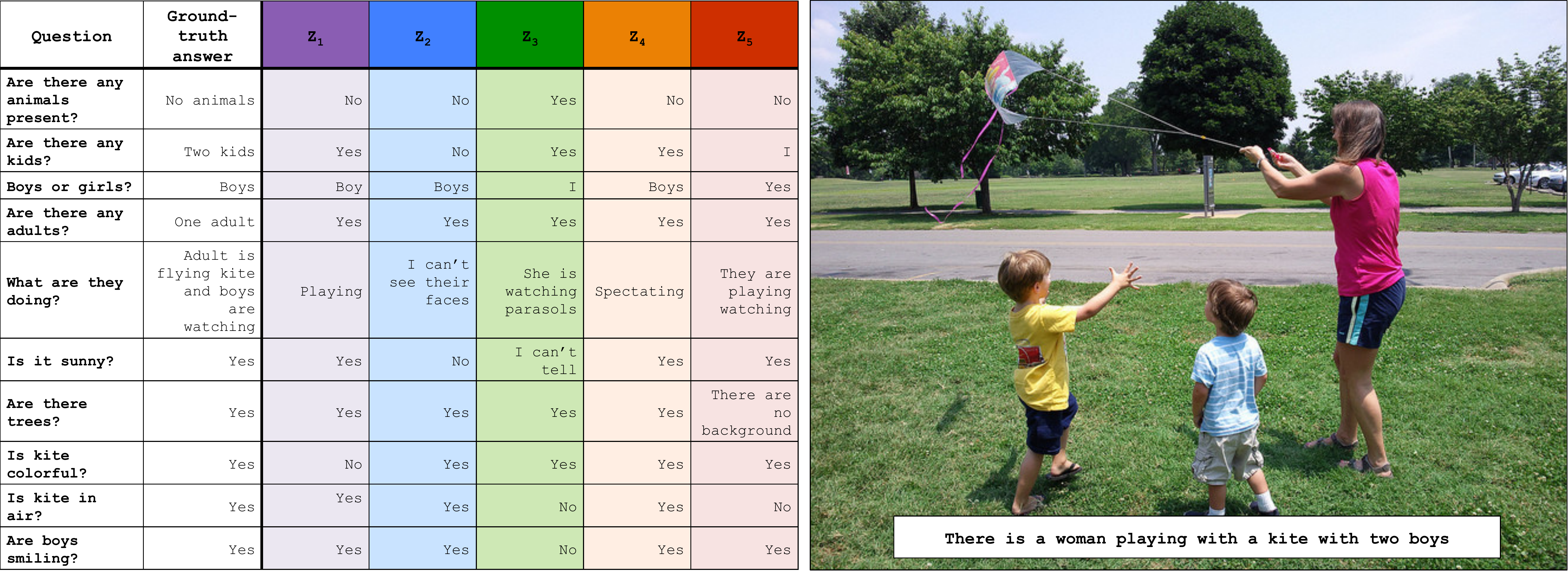}
  \end{tabular}
  \caption{Examples of diverse answer generations from the \modx{} model for the \gls{1VD} task.}
  \label{fig:qual_gen2_supp1}
\end{figure*}

\begin{figure*}[!h]
  \centering
  \begin{tabular}{@{}c@{}}
    \includegraphics[width=0.92\linewidth]{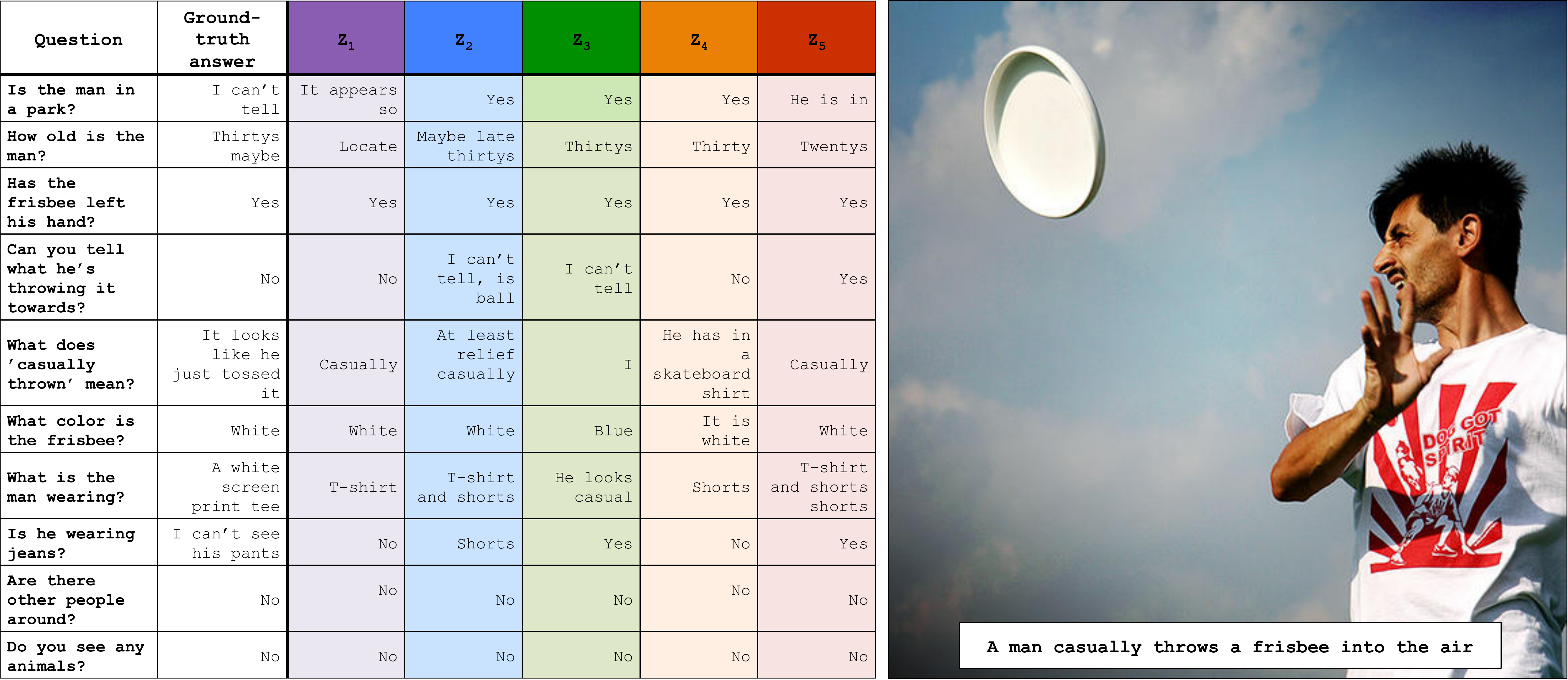}\\
    \includegraphics[width=0.92\linewidth]{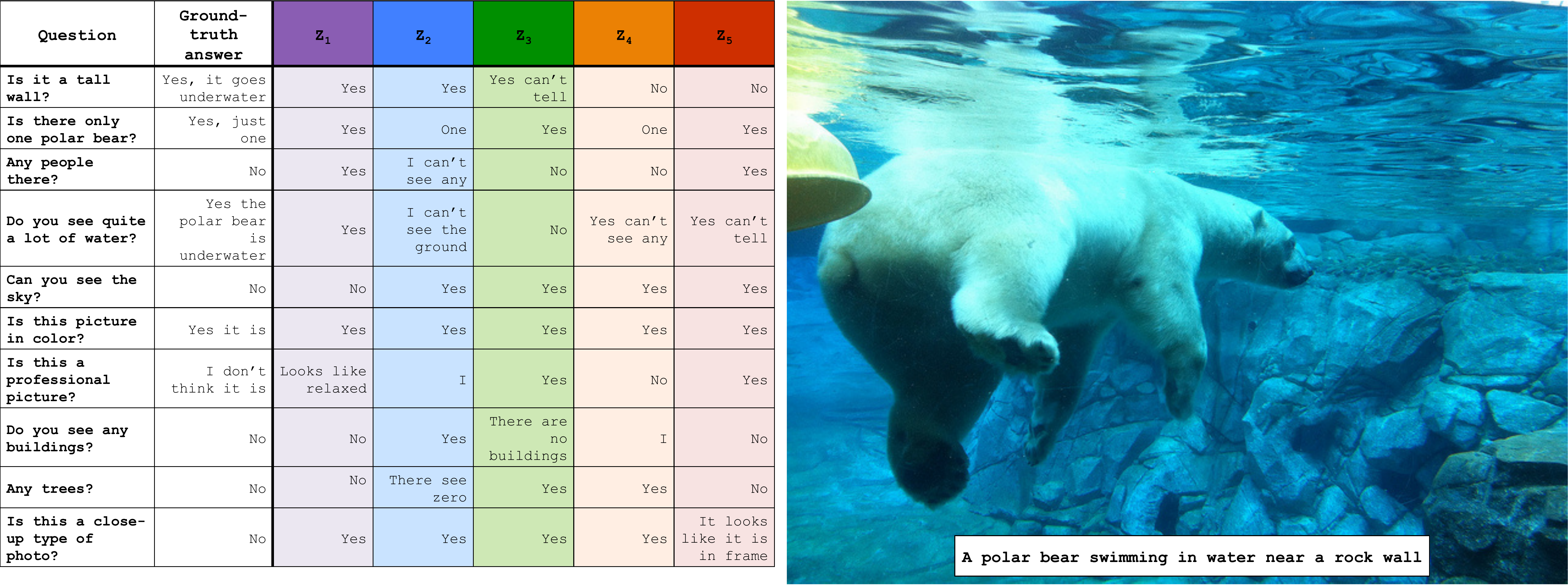}\\
    \includegraphics[width=0.92\linewidth]{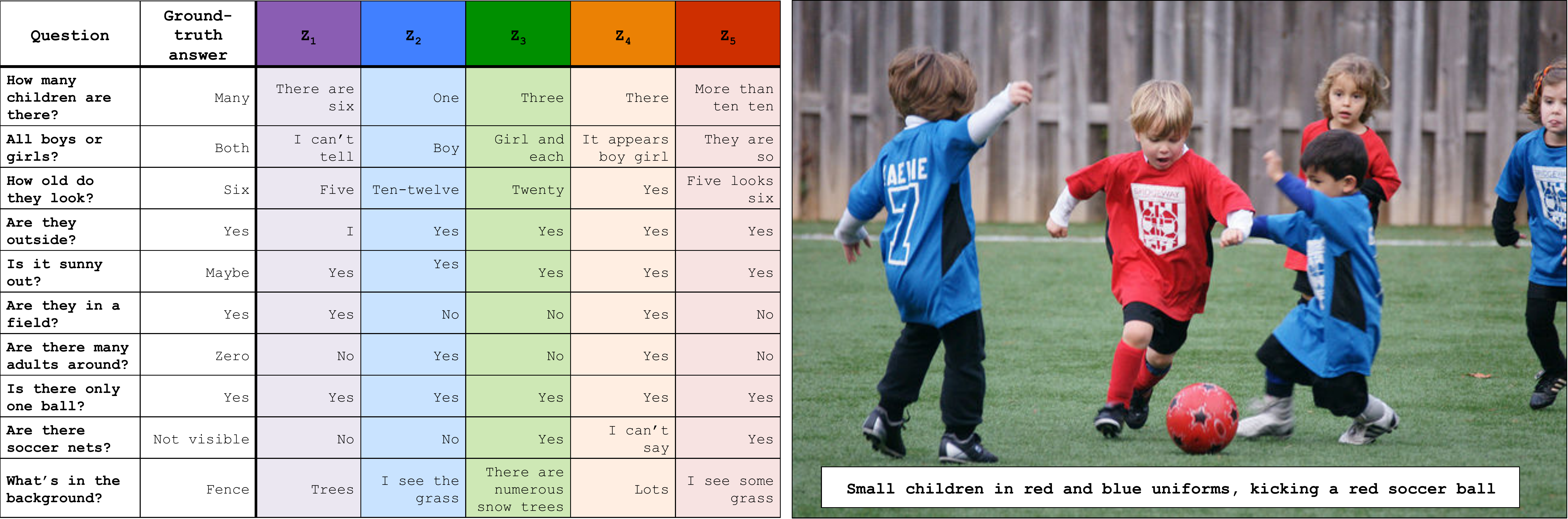}\\[15ex]
  \end{tabular}
  \caption{Examples of diverse answer generations from the \modx{} model for the \gls{1VD} task -- continued.}
  \label{fig:qual_gen2_supp2}
\end{figure*}

\begin{figure*}[!h]
  \centering
  \begin{tabular}{@{}c@{}}
  \includegraphics[width=0.98\linewidth]{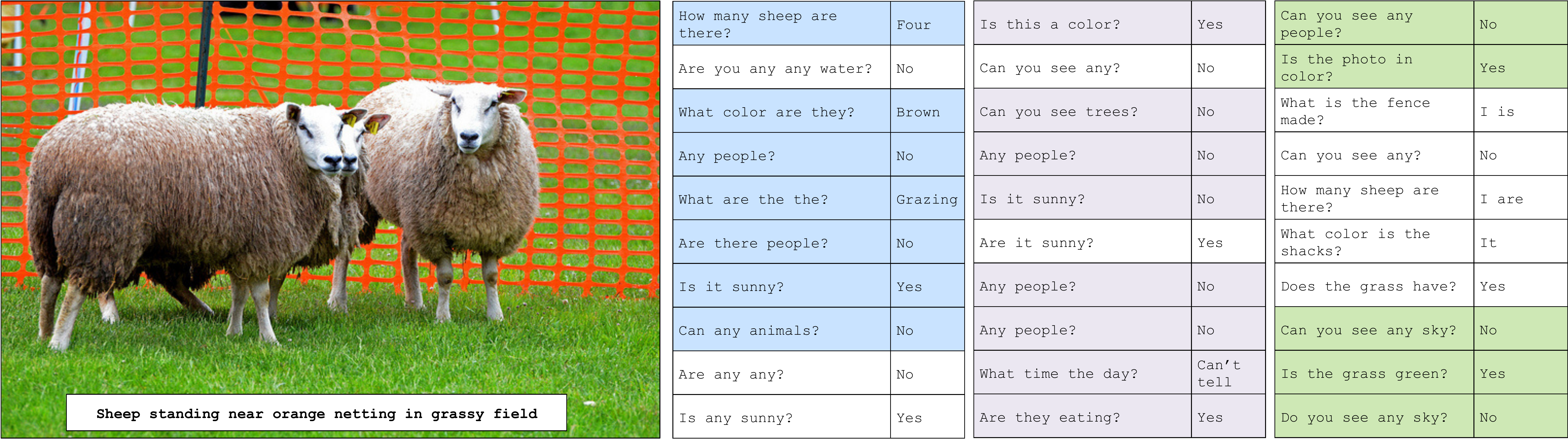}\\
  \includegraphics[width=0.98\linewidth]{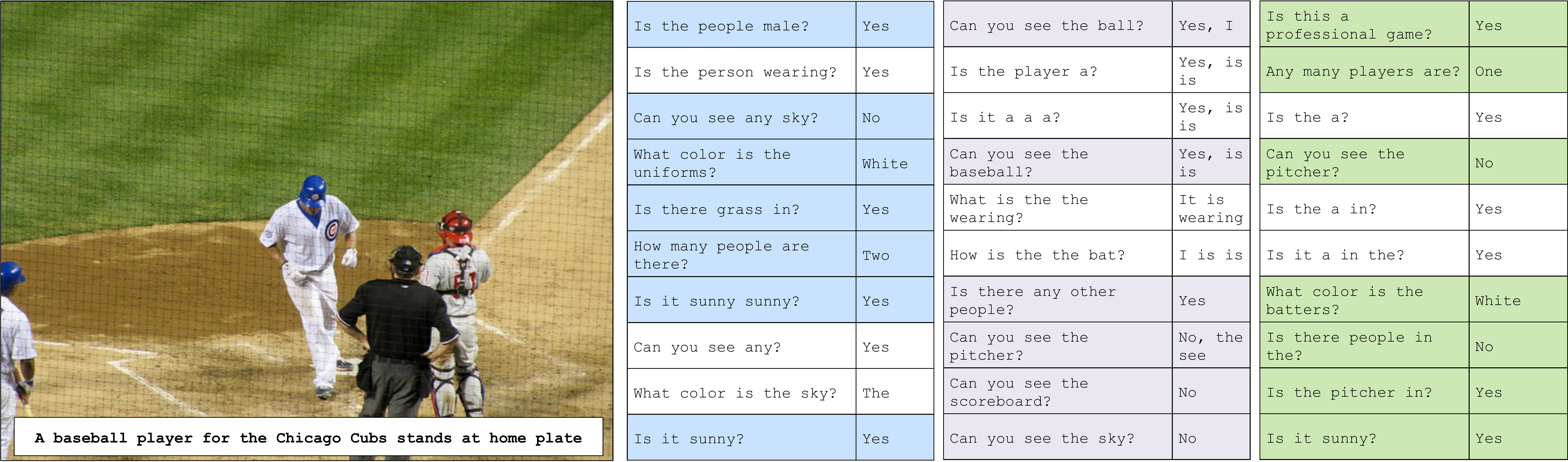}\\
  \includegraphics[width=0.98\linewidth]{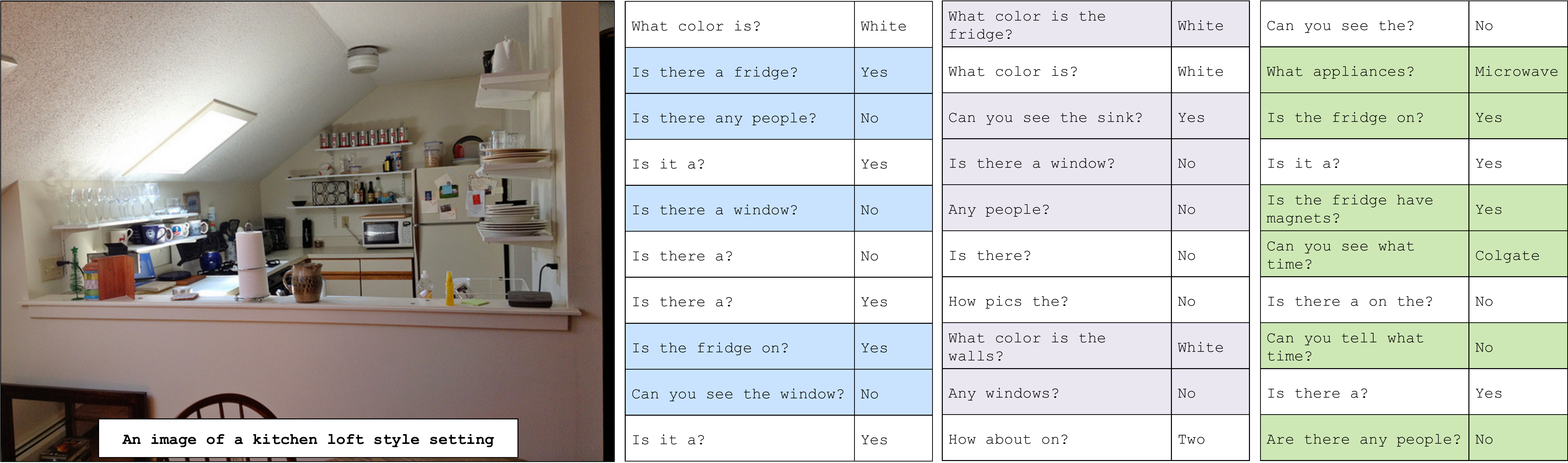}\\
  \includegraphics[width=0.98\linewidth]{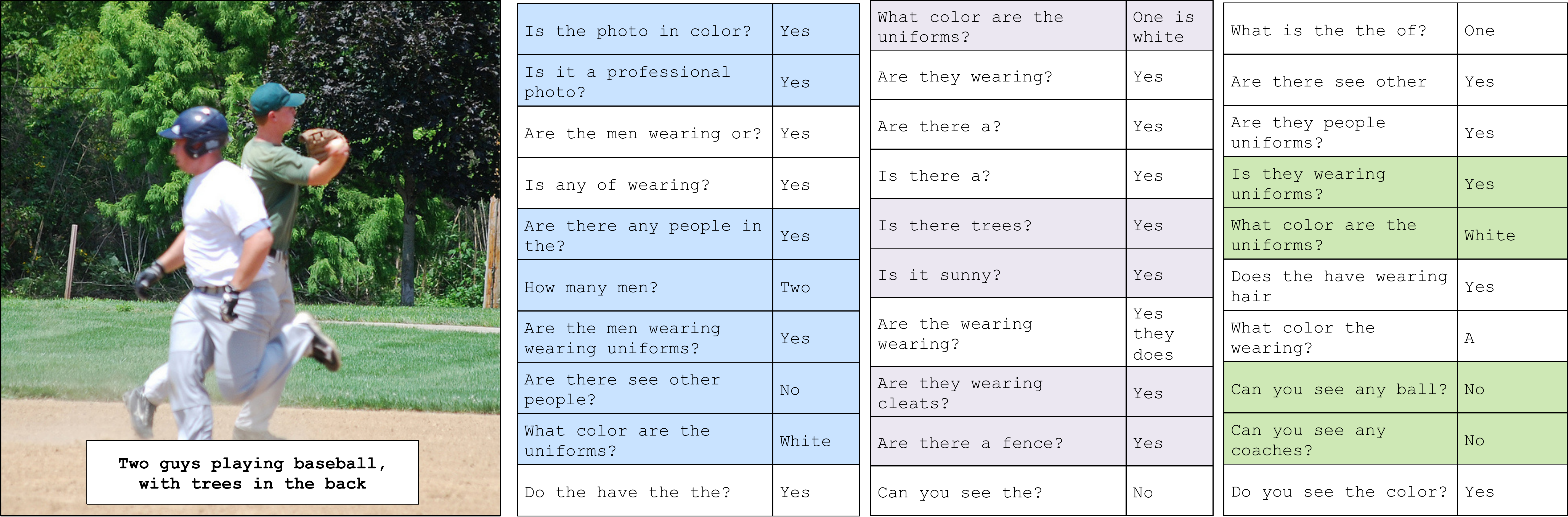}
  \end{tabular}
  \caption{Diverse two-way dialogue generations from the \modzar10{} model (block evaluation) for the \gls{2VD} task.}
  \label{fig:qual_gen1_supp1}
\end{figure*}

\begin{figure*}[!h]
  \centering
  \begin{tabular}{@{}c@{}}
  \includegraphics[width=0.98\linewidth]{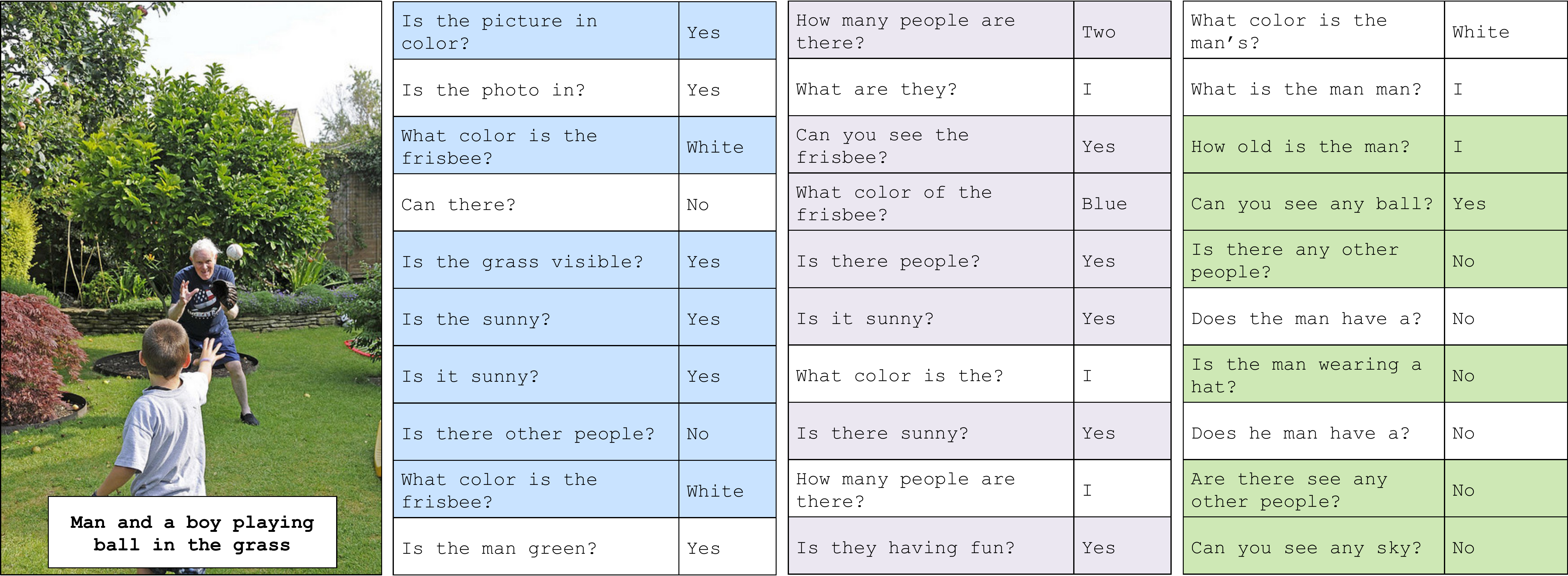}\\
  \includegraphics[width=0.98\linewidth]{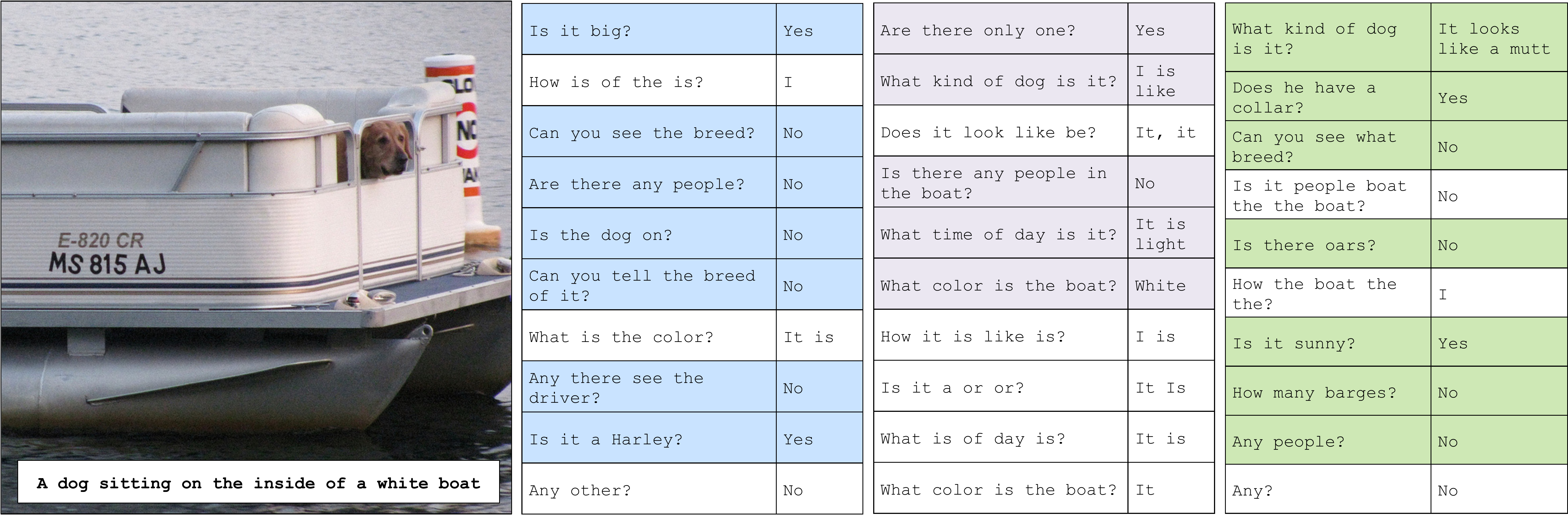}\\
  \includegraphics[width=0.98\linewidth]{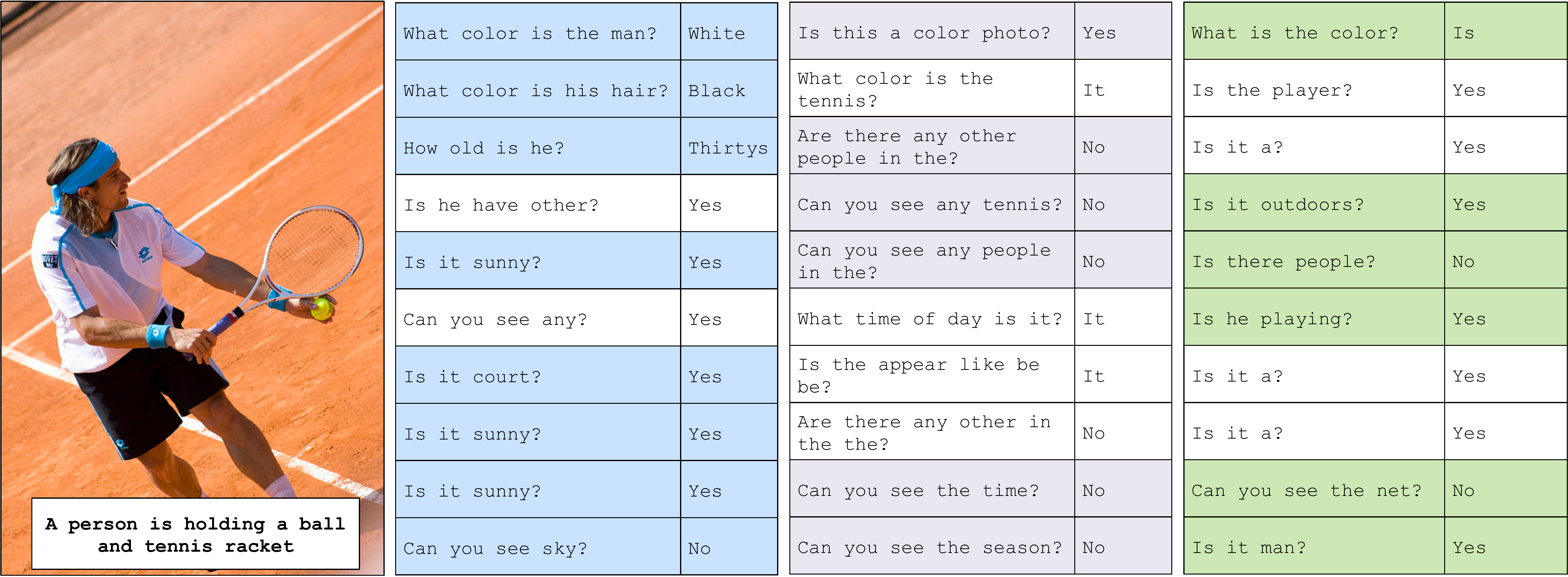}\\
  \end{tabular}
  \caption{Diverse two-way dialogue generations from the \modzar10{} model (block evaluations) for the \gls{2VD} task -- continued}
  \label{fig:qual_gen1_supp2}
\end{figure*}

\vspace*{-1ex}
We present additional qualitative results for the \modx{} model in \cref{fig:qual_gen2_supp1,fig:qual_gen2_supp2} (\gls{1VD} task) and for the \modzar10{} model (under the block evaluation setting) in \cref{fig:qual_gen1_supp1,fig:qual_gen1_supp2} (\gls{2VD} task).
Note that for both, different colours indicate generations (\(\hat{\ans}_t\) for \modx{} and \(\hat{\dial}\) for \modz/\modzar{}) from different samples of \(\z\). In \cref{fig:qual_gen1_supp1,fig:qual_gen1_supp2}, whole generated dialogue blocks are shown with coloured sections indicating subsets exhibiting coherent question-answering and white sections indicating subsets that are not entirely coherent.

\end{document}